\def\year{2019}\relax
\documentclass[letterpaper]{article} 
\usepackage{aaai19}  
\usepackage{times}  
\usepackage{helvet}  
\usepackage{courier}  
\usepackage{url}  
\usepackage{graphicx}  
\usepackage{bm}
\usepackage{amsmath}
\usepackage{amssymb}

\def\S{\mathcal{S}}
\def\A{\mathcal{A}}

\begin{document}
\title{Multi-task Deep Reinforcement Learning with PopArt}
\author{
   Matteo Hessel \And
   Hubert Soyer \And
   Lasse Espeholt \AND
   Wojciech Czarnecki \And
   Simon Schmitt \And
   Hado van Hasselt
}
\maketitle

\begin{abstract}
The reinforcement learning community has made great strides in designing algorithms capable of exceeding human performance on specific tasks. These algorithms are mostly trained one task at the time, each new task requiring to train a brand new agent instance. This means the learning algorithm is general, but each solution is not; each agent can only solve the one task it was trained on. In this work, we study the problem of learning to master not one but multiple sequential-decision tasks at once. A general issue in multi-task learning is that a balance must be found between the needs of multiple tasks competing for the limited resources of a single learning system.
Many learning algorithms can get distracted by certain tasks in the set of tasks to solve.  Such tasks appear more salient to the learning process, for instance because of the density or magnitude of the in-task rewards. This causes the algorithm to focus on those salient tasks at the expense of generality.
We propose to automatically adapt the contribution of each task to the agent's updates, so that all tasks have a similar impact on the learning dynamics. This resulted in state of the art performance on learning to play all games in a set of 57 diverse Atari games. Excitingly, our method learned a single trained policy - with a single set of weights - that exceeds median human performance. To our knowledge, this was the first time a single agent surpassed human-level performance on this multi-task domain. The same approach also demonstrated state of the art performance on a set of 30 tasks in the 3D reinforcement learning platform DeepMind Lab.
\end{abstract}

\section{Introduction}

In recent years, the field of deep reinforcement learning (RL) has enjoyed many successes. Deep RL agents have been applied to board games such as Go \cite{Silver_2016} and chess \cite{AlphaZero}, continuous control \cite{Lillicrap2016,duan2016benchmarking}, classic video-games such as Atari \cite{Mnih2015,Rainbow,Reactor,SchulmanTRPO,SchulmanPPO,OptionCritic}, and 3D first person environments \cite{Mnih2016,UNREAL}. While the results are impressive, they were achieved on one task at the time, each task requiring to train a new agent instance from scratch. 

Multi-task and transfer learning remain important open problems in deep RL. There are at least four different strains of multi-task reinforcement learning that have been explored in the literature: off-policy learning of many predictions about the same stream of experience \cite{Schmidhuber90,Horde,UNREAL}, continual learning in a sequence of tasks \cite{RingContinualPhD94,thrun1996learning,thrun2012explanation,Progressive}, distillation of task-specific experts into a single shared model \cite{ActorMimic,Distillation,Kickstarting,Distral}, and parallel learning of multiple tasks at once \cite{SharmaMultiTask,caruana1998multitask}. We will focus on the latter.

Parallel multi-task learning has recently achieved remarkable success in enabling a single system to learn a large number of diverse tasks. The Importance Weighted Actor-Learner Architecture, henceforth IMPALA \cite{IMPALA}, achieved a 59.7\% median human normalised score across 57 Atari games, and a 49.4\% mean human normalised score across 30 DeepMind Lab levels. These results are state of the art for multi-task RL, but they are far from the human-level performance demonstrated by deep RL agents on the same domains, when trained on each task individually.

Part of why multi-task learning is much harder than single task learning is that a balance must be found between the needs of multiple tasks, that compete for the limited resources of a single learning system (for instance, for its limited representation capacity). We observed that the naive transposition of common RL algorithms to the multi-task setting may not perform well in this respect. More specifically, the \textit{saliency} of a task for the agent increases with the scale of the \textit{returns} observed in that task, and these may differ arbitrarily across tasks. This affects value-based algorithms such as Q-learning \cite{Watkins:1989}, as well as policy-based algorithms such as REINFORCE \cite{Williams1992}.

The problem of scaling individual rewards appropriately is not novel, and has often been addressed through reward clipping \cite{Mnih2015}. This heuristic changes the agent's objective, e.g., if all rewards are non-negative the algorithm optimises \emph{frequency} of rewards rather than their cumulative sum. If the two objectives are sufficiently well aligned, clipping can be effective. However, the scale of returns also depends on the rewards' sparsity. This implies that, even with reward clipping, in a multi-task setting the magnitude of updates can still differ significantly between tasks, causing some tasks to have a larger impact on the learning dynamics than other equally important ones. 

Note that both the sparsity and the magnitude of rewards collected in an environment are inherently \textit{non-stationary}, because the agent is learning to actively maximise the total amount of rewards it can collect. These non-stationary learning dynamics make it impossible to normalise the learning updates a priori, even if we would be willing to pour significant domain knowledge into the design of the algorithm.

To summarise, in IMPALA the magnitude of updates resulting from experience gathered in each environment depends on: 1) the \textit{scale} of rewards, 2) the \textit{sparsity} of rewards, 3) the \textit{competence} of the agent. In this paper we use PopArt normalisation \cite{PopNIPS} to derive an actor-critic update invariant to these factors, enabling large performance improvements in parallel multi-task agents. We demonstrated this on the Atari-57 benchmark, where a single agent achieved a median normalised score of 110\% and on DmLab-30, where it achieved a mean score of 72.8\%. 

\section{Background}

Reinforcement learning (RL) is a framework for learning and decision-making under uncertainty \cite{SB2018}. A learning system - the \textit{agent} - must learn to interact with the \textit{environment} it is embedded in, so as to maximise a scalar \textit{reward}  signal. The RL problem is often formalised as a Markov decision process \cite{bellmanMDP}: a tuple $(\S, \A, p, \gamma)$, where $\S, \A$ are finite sets of \textit{states} and \textit{actions}, $p$ denotes the dynamics, such that $p(r, s' \mid s, a)$ is the probability of observing reward $r$ and state $s'$ when executing action $a$ in state $s$, and $\gamma \in [0,1]$ discounts future rewards. The \textit{policy} maps states $s \in \S$ to probability distributions over actions $\pi(A|S=s)$, thus specifying the behaviour of the agent. The \textit{return} $G_t = R_{t+1} + \gamma R_{t+2} + \ldots$ is the $\gamma$-discounted sum of rewards collected by an agent from state $S_t$ onward under policy $\pi$. We define \textit{action values} and \textit{state values} as $q^\pi(s, a) = \mathbb{E}_{\pi}[ G_t \mid S_t = s, A_t=a]$ and $v^\pi(s) = \mathbb{E}_{\pi}[ G_t \mid S_t = s]$, respectively. The agent's objective is to find a policy to maximise such values.

In multi-task reinforcement learning, a single agent must learn to master N different environments $T = \{D_i = (\S_i, \A_i, p_i, \gamma)\}_{i=1}^N$, each with its own distinct dynamics \cite{Brunskill2013SampleCO}. Particularly interesting is the case in which the action space and transition dynamics are at least partially shared. For instance, the environments might follow the same \textit{physical} rules, while the set of interconnected states and obtainable rewards differ. We may formalise this as a single larger MDP, whose state space is $\S = \{\{(s_j, i)\}_{s_j \in \S_i}\}_{i=1}^N$. The task index $i$ may be latent, or may be exposed to the agent's policy. In this paper, we use the task index at training time, for the value estimates used to compute the policy updates, but not at testing time: our algorithm will return a single general policy $\pi(A|S)$ which is only function of the individual environment's state $S$ and not conditioned directly on task index $i$. This is more challenging than the standard multi-task learning setup, which typically allows conditioning the model on the task index even at evaluation \cite{MultiPontil,MultiNLP}, because our agents will need to infer what task to solve purely from the stream of raw observations and/or early rewards in the episode.

\subsection{Actor-critic}
In our experiments, we use an actor-critic algorithm to learn a policy $\pi_{\eta}(A|S)$ and a value estimate $v_{\theta}(s)$, which are both outputs of a deep neural network.  We update the agent's policy by using REINFORCE-style stochastic gradient $(G_t - v_{\theta}(S_t)) \nabla_{\eta} \log \pi(A_t | S_t)$ \cite{Williams1992}, where $v_{\theta}(S_t)$ is used as a baseline to reduce variance. In addition we use a multi-step return $G^v_t$ that bootstraps on the value estimates after a limited number of transitions, both to reduce variance further and to allow us to update the policy before $G_t$ fully resolves at the end of an episode. The value function $v_{\theta}(S)$ is instead updated to minimise the squared loss with respect to the (truncated and bootstrapped) return:
\begin{equation}\label{v-eq}
\Delta \theta \propto -\nabla_{\theta}(G_t^v - v_{\theta}(S_{t}))^2 = (G_t^v - v_{\theta}(S_t)) \nabla_{\theta} v_{\theta}(S_t) \,
\end{equation}
\begin{equation}\label{pi-eq}
\Delta \eta \propto (G_t^\pi - v_{\theta}(S_t)) \nabla_{\eta} \log( \pi_{\eta}(A_t|S_t)) \,,
\end{equation}
where $G_t^v$ and $G_t^\pi$ are stochastic estimates of $v^\pi(S_t)$ and $q^\pi(S_t, A_t)$, respectively.
Note how both updates depend linearly on the scale of returns, which, as previously argued, depend on scale/sparsity of rewards, and agent's competence. 

\subsection{Efficient multi-task learning in simulation}
We use the IMPALA agent architecture \cite{IMPALA}, proposed for reinforcement learning in simulated environments. In IMPALA the agent is distributed across multiple threads, processes or machines. Several actors run on CPU generating rollouts of experience, consisting of a fixed number of interactions (100 in our experiments) with their own copy of the environment, and then enqueue the rollouts in a shared queue. Actors receive the latest copy of the network's parameters from the learner before each rollout. A single GPU learner processes rollouts from all actors, in batches, and updates a deep network. The network is a deep convolutional ResNet \cite{He:2015}, followed by a LSTM recurrent layer \cite{Hochreiter:1997}. Policy and values are all linear functions of the LSTM's output.

Despite the large network used for estimating policy $\pi_{\eta}$ and values $v_{\theta}$, the decoupled nature of the agent enables to process data very efficiently: in the order of hundreds of thousands frames per second \cite{IMPALA}. The setup easily supports the multi-task setting by simply assigning different environments to each of the actors and then running the single policy $\pi(S|A)$ on each of them. The data in the queue can also be easily labelled with the task \textit{id}, if useful at training time. Note that an efficient implementation of IMPALA is available open-source \footnote{www.github.com/deepmind/scalable\_agent}, and that, while we use this agent for our experiments, our approach can be applied to other data parallel multi-task agents (e.g. A3C).

\subsection{Off-policy corrections} 
Because we use a distributed queue-based learning setup, the data consumed by the learning algorithm might be slightly off-policy, as the policy parameters change between acting and learning.  We can use importance sampling corrections $\rho_t = \pi(A_t | S_t) / \mu(A_t | S_t)$ to compensate for this \cite{Precup:2000}.  In particular, we can write the $n$-step return as
$
G_t = R_{t+1} + \gamma R_{t+2} + \ldots + \gamma^n v(S_{t+n}) = v(S_t) + \sum_{k=t}^{t+n-1} \gamma^{k-t} \delta_k,
$
where $\delta_t = R_{t+1} + \gamma v(S_{t+1}) - v(S_t)$, and then apply appropriate importance sampling corrections to each error term \cite{Sutton:2014} to get
$
G_t = v(S_t) + \sum_{k=t}^{t+n-1} \gamma^{k-t} (\prod_{i=t}^{k} \rho_i ) \delta_k .
$
This is unbiased, but has high variance.  To reduce variance, we can further clip most of the importance-sampling ratios, e.g., as $c_t = \min(1, \rho_t)$.  This leads to the $v$-trace return \cite{IMPALA}
\begin{equation}
G_t^{v-trace} = v(S_t) + \sum_{k=t}^{t+n-1} \gamma^{k-t} \left(\prod_{i=t}^{k} c_i\right) \delta_k
\end{equation}

A very similar target was proposed for the ABQ($\zeta$) algorithm \cite{Mahmood:2017}, where the product $\rho_t \lambda_t$ was considered and then the trace parameter $\lambda_t$ was chosen to be adaptive to lead to exactly the same behaviour that $c_t = \rho_t \lambda_t = \min(1, \rho_t)$.  This shows that this form of clipping does not impair the validity of the off-policy corrections, in the same sense that bootstrapping in general does not change the semantics of a return.
The returns used by the value and policy updates defined in Equation \ref{v-eq} and \ref{pi-eq} are then
\begin{equation}
G^v_t \ = G_t^{v-trace}
\quad \mbox{and} \quad
G^\pi_t \ = R_{t+1} + \gamma G_{t+1}^{v\text{-trace}}.
\end{equation}

This is the same algorithm as used by Espeholt et al. (2018) in the experiments on the IMPALA architecture.

\section{Adaptive normalisation}

In this section we use PopArt normalisation \cite{PopNIPS}, which was introduced for value-based RL, to derive a scale invariant algorithm for actor-critic agents. For simplicity, we first consider the single-task setting, then we extend it to the multi-task setting (the focus of this work).

\subsection{Scale invariant updates}

In order to normalise both baseline and policy gradient updates, we first parameterise the value estimate $v_{\mu, \sigma, \theta}(S)$ as the linear transformation of a suitably \textit{normalised} value prediction $n_{\theta}(S)$. We further assume that the normalised value prediction is itself the output of a linear function, for instance the last fully connected layer of a deep neural net:
\begin{equation}
v_{\mu, \sigma, \theta}(s) = \sigma \cdot n_{\theta}(s) + \mu  = \sigma \cdot (\underbrace{\bm{w}^{\top}f_{\theta \backslash \{w, b\}}(s) + b}_{\mbox{$= n_{\theta}(s)$}}) + \mu \,.
\label{PopSingle}
\end{equation}
As proposed by van Hasselt et al., $\mu$ and $\sigma$ can be updated so as to track mean and standard deviation of the values. First and second moments of can be estimated online as
\begin{equation}
\mu_t = (1-\beta) \mu_{t-1} + \beta G_t^v, \qquad
\nu_t = (1-\beta) \nu_{t-1} + \beta (G_t^v)^2, \qquad
\label{stats-rule}
\end{equation}
and then used to derive the estimated standard deviation as $\sigma_t = \sqrt{\nu_t - \mu_t^2}$. Note that the fixed \textit{decay} rate $\beta$ determines the horizon used to compute the statistics. We can then use the normalised value estimate $n_{\theta}(S)$ and the statistics $\mu$\ and $\sigma$ to normalise the actor-critic loss, both in its value and policy component; this results in the scale-invariant updates:
\begin{equation}
\label{popart-updates-1}
\Delta \theta \propto \left(\frac{G_t^v -\mu}{\sigma} - n_{\theta}(S_t)\right) \nabla_{\theta} n_{\theta}(S_t) \,,
\end{equation}
\vspace{-10pt}
\begin{equation}
\label{popart-updates-2}
\Delta \eta \propto  \left(\frac{G_t^\pi - \mu}{\sigma} - n_{\theta}(S_t) \right) \nabla_{\eta} \log \pi_{\eta}(A_{t}|S_{t}) \,.
\end{equation}
If we optimise the new objective naively, we are at risk of making the problem harder: the normalised targets for values are non-stationary, since they depend on statistics $\mu$ and $\sigma$. The PopArt normalisation algorithm prevents this, by updating the last layer of the normalised value network to preserve unnormalised value estimates $v_{\mu, \sigma, \theta}$, under any change in the statistics $\mu \rightarrow \mu'$ and $\sigma \rightarrow \sigma'$:
\begin{equation}
\bm{w}' = \frac{\sigma}{\sigma'}\bm{w} \,,
\qquad
b' = \frac{\sigma b + \mu - \mu'}{\sigma'} \,.
\end{equation}
This extends PopArt's scale-invariant updates to the actor-critic setting, and can help to make tuning hyperparameters easier, but it is not sufficient to tackle the challenging multi-task RL setting that we are interested in this paper. For this, a single pair of normalisation statistics is not sufficient.

\subsection{Scale invariant updates for multi-task learning}

Let $D_i$ be an environment in some finite set $T=\{D_i\}_{i=1}^N$, and let $\pi(S|A)$ be a task-agnostic policy, that takes a state $S$ from any of the environments $D_i$, and maps it to a probability distribution onto the shared action space $\A$. Consider now a multi-task value function $\mathbf{v}(S)$ with N outputs, one for each task.
We can use for $\mathbf{v}$ the same parametrisation as in Equation \ref{PopSingle}, but with vectors of statistics $\bm{\mu}, \bm{\sigma} \in \mathbb{R}^N$, and a vector-valued function $\mathbf{n}_{\theta}(s) = (n^1_{\theta}(s), \ldots, n^N_{\theta}(s))^{\top}$

\begin{equation}
\mathbf{v}_{\bm{\mu}, \bm{\sigma}, \theta}(S) = 
\bm{\sigma} \odot \mathbf{n}_{\theta}(S) + \bm{\mu} =
\bm{\sigma} \odot (\mathbf{W} \mathbf{f}_{\theta \backslash \{\mathbf{W}, \mathbf{b}\}}(S) + \mathbf{b}) + \bm{\mu}
\end{equation}
where $\mathbf{W}$ and $\mathbf{b}$ denote the parameters of the last fully connected layer in $\mathbf{n}_{\theta}(s)$. Given a rollout $\{S_{i,k}, A_k, R_{i,k}\}^{t+n}_{k=t}$, generated under task-agnostic policy $\pi_{\eta}(A|S)$ in environment $D_i$, we can adapt the updates in Equation \ref{popart-updates-1} and \ref{popart-updates-2} to provide scale invariant updates also in the multi-task setting:
\begin{align}
\label{popart-updates-multi}
\Delta \theta & \propto \left(\frac{G^{v, i}_{t} -\mu_i}{\sigma_i} - n^i_{\theta}(S_t)\right) \nabla_{\theta} n^i_{\theta}(S_t) \,,
\\
\Delta \eta & \propto  \left(\frac{G^{\pi, i}_{t} - \mu_i}{\sigma_i} - n^i_{\theta}(S_t) \right) \nabla_{\eta} \log \pi_{\eta}(A_{t}|S_{t}) \,.
\end{align}
Where the targets $G^{\cdot, i}_t$ use the value estimates for environment $D_i$ for bootstrapping. For each rollout, only the $i^{\text{th}}$ head in the value net is updated, while the same policy network is updated irrespectively of the task, using the appropriate rescaling for updates to parameters $\eta$. As in the single-task case, when updating the statistics $\bm{\mu}$ and $\bm{\sigma}$ we also need to update $\mathbf{W}$ and $\mathbf{b}$ to preserve unnormalised outputs,
\begin{align}\label{popupdate}
\mathbf{w}_i' &= \frac{\sigma_i}{\sigma_i'} \mathbf{w}_i \,,
&
b_i' = \frac{\sigma_i b_i + \mu_i - \mu_i'}{\sigma_i'} \,,
\end{align}
where $\mathbf{w}_i$ is the $i^{\text{th}}$ row of matrix $\mathbf{W}$, and $\mu_i, \sigma_i, b_i$ are the $i^{\text{th}}$ elements of the corresponding parameter vectors.
Note that in all updates only the values, but not the policy, are conditioned on the task index, which ensures that the resulting agent can then be run in a fully task agnostic way, since values are only used to reduce the variance of the policy updates at training time but not needed for action selection.

\begin{figure*}[ht]
\renewcommand{\figurename}{Table}
\centering
\caption{\textbf{Summary of results}: aggregate scores for IMPALA and PopArt-IMPALA. We report median human normalised score for Atari-57, and mean capped human normalised score for DmLab-30. In Atari, Random and Human refer to whether the trained agent is evaluated with random or human starts. In DmLab-30 the test score includes evaluation on the held-out levels. \label{summary-score-table}}
\vspace{5pt}
\bgroup
\def\arraystretch{1.5}
\small{
\begin{tabular}{ l  r  r c r  r  c r  r }

  & \multicolumn{2}{c}{Atari-57}
  &
  & \multicolumn{2}{c}{Atari-57 (unclipped)}
  &
  & \multicolumn{2}{c}{DmLab-30} \\ 
 \cline{2-3}  \cline{5-6}   \cline{8-9} 
 Agent   & Random &  Human &&  Random  & Human && Train & Test \\

 IMPALA        &  $59.7\%$ & $28.5\%$  &  &  $0.3\%$ & $1.0\%$     &  & $60.6\%$ & $58.4\%$  \\
 PopArt-IMPALA &  $110.7\%$ & $101.5\%$  & & $107.0\%$ & $93.7\%$&  & $73.5\%$ & $72.8\%$  \\ 

\end{tabular}}
\egroup
\vspace{9pt}
\end{figure*}

\newpage
\section{Experiments}\label{exp}
\vspace{10pt}

We evaluated our approach in two challenging multi-task benchmarks, Atari-57 and DmLab-30, based on Atari and DeepMind Lab respectively, and introduced by Espeholt et al. We also consider a new benchmark, consisting of the same games as Atari-57, but with the original unclipped rewards. We demonstrate state of the art performance on all benchmarks. To aggregate scores across many tasks, we normalise the scores on each task based on the scores of a human player and of a random agent on that same task \cite{van2016deep}. All experiments use population-based training (PBT) to tune hyperparameters \cite{Jaderberg:2017}. As in Espeholt et al., we report learning curves as function of the number of frames processed by one instance of the tuning population, summed across tasks.

\subsection{Domains}

Atari-57 is a collection of 57 classic Atari 2600 games. The ALE \cite{bellemare2013arcade}, exposes them as RL environments. Most prior work has focused on training agents for individual games \cite{Mnih2015,Rainbow,Reactor,SchulmanTRPO,SchulmanPPO,OptionCritic}. Multi-task learning on this platform has not been as successful due to large number of environments, inconsistent dynamics and very different reward structure. Prior work on multi-task RL in the ALE has therefore focused on smaller subsets of games \cite{Distillation,SharmaMultiTask}. Atari has a particularly diverse reward structure. Consequently, it is a perfect domain to fully assess how well can our agents deal with extreme differences in the scale of returns. Thus, we train all agents both with and without reward clipping, to compare performance degradation as returns get more diverse in the unclipped version of the environment. In both cases, at the end of training, we test agents both with \textit{random-starts} \cite{Mnih2015} and \textit{human-starts} \cite{Nair2015}; aggregate results are reported in Table \ref{summary-score-table} accordingly.

DmLab-30 is a collection of 30 visually rich, partially observable RL environments \cite{DmLab}. This benchmark has strong internal consistency (all levels are played with a first person camera in a 3D environment with consistent dynamics). Howevere, the tasks themselves are quite diverse, and were designed to test distinct skills in RL agents: among these navigation, memory, planning, laser-tagging, and language grounding. The levels can also differ visually in non-trivial ways, as they include both natural environments and maze-like levels. Two levels (\texttt{rooms\_collect\_good\_objects} and \texttt{rooms\_exploit\_deferred\_effects}) have held out \textit{test} versions, therefore Table \ref{summary-score-table} reports both train and test aggregate scores. We observed that the original IMPALA agent suffers from an artificial bottleneck in performance, due to the fact that some of the tasks cannot be solved with the action set available to the agent. As first step, we thus fix this issue by equipping it with a larger action set, resulting in a stronger IMPALA baseline than reported in the original paper. We also run multiple independent PBT experiments, to assess the variability of results across multiple replications. 

\subsection{Atari-57 results}

Figures \ref{fig:atari-curve-clipped} and \ref{fig:atari-curve-unclipped} show the median human normalised performance across the entire set of 57 Atari games in the ALE, when training agent with and without reward clipping, respectively. The curves are plotted as function of the total number of frames seen by each agent.

PopArt-IMPALA (orange line) achieves a median performance of 110\% with reward clipping and a median performance of 101\% in the unclipped version of Atari-57. Recall that here we are measuring the median performance of a single trained agent across all games, rather than the median over the performance of a set of individually trained agents as it has been more common in the Atari domain. To our knowledge, both agents are the first to surpass median human performance across the entire set of 57 Atari games.

The IMPALA agent (blue line) performs much worse. The baseline barely reaches $60\%$ with reward clipping, and the median performance is close to 0\% in the unclipped setup. The large decrease in the performance of the baseline IMPALA agent once clipping is removed is in stark contrast with what we observed for PopArt-IMPALA, that achieved almost the same performance in the two training regimes. 

Since the level-specific value predictions used by multi-task PopArt effectively increase the capacity of the network, we also ran an additional experiment to disentangle the contribution of the increased network capacity from the contribution of the adaptive normalisation. For this purpose, we train a second baseline, that uses level specific value predictions, but does not use PopArt to adaptively normalise the learning updates. The experiments show that such MultiHead-IMPALA agent (pink line) actually performs slightly worse than the original IMPALA both with and without clipping, confirming that the performance boost of PopArt-IMPALA is indeed due to the adaptive rescaling.

We highlight that in our experiments a single instance of multi-task PopArt-IMPALA has processed the same amount of frames as a collection of 57 expert DQN agents ($57 \times 200\,\text{M} = 1.14 \times 10^{10}$), while achieving better performance. Despite the large CPU requirements, on a cloud service, training multi-task PopArt-IMPALA can also be competitive in terms of costs, since it exceeds the performance of a vanilla-DQN in just 2.5 days, with a smaller GPU footprint.

\setcounter{figure}{0}
\renewcommand{\figurename}{Figure}

\begin{figure}[t]
\centering
\includegraphics[width=\linewidth]{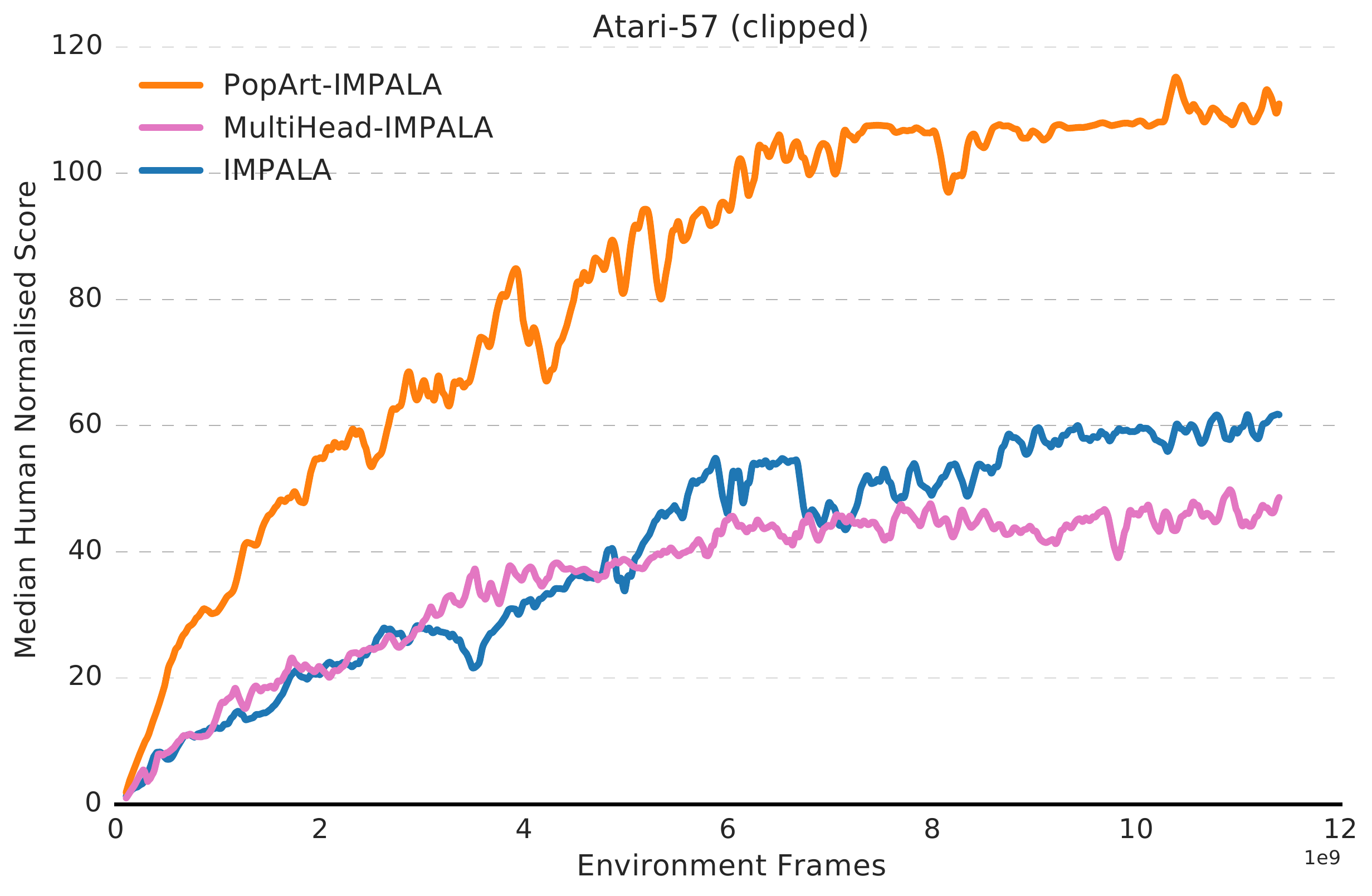}
\caption{\textbf{Atari-57 (reward clipping)}. Median human normalised score across all Atari levels, as function of the total number of frames seen by the agents across all levels. We compare PopArt-IMPALA to IMPALA and to an additional baseline, MultiHead-IMPALA, that uses task-specific value predictions but no adaptive normalisation. All three agent are trained with the clipped reward scheme.}
\label{fig:atari-curve-clipped}
\end{figure}

\begin{figure}[t]
\centering
\includegraphics[width=\linewidth]{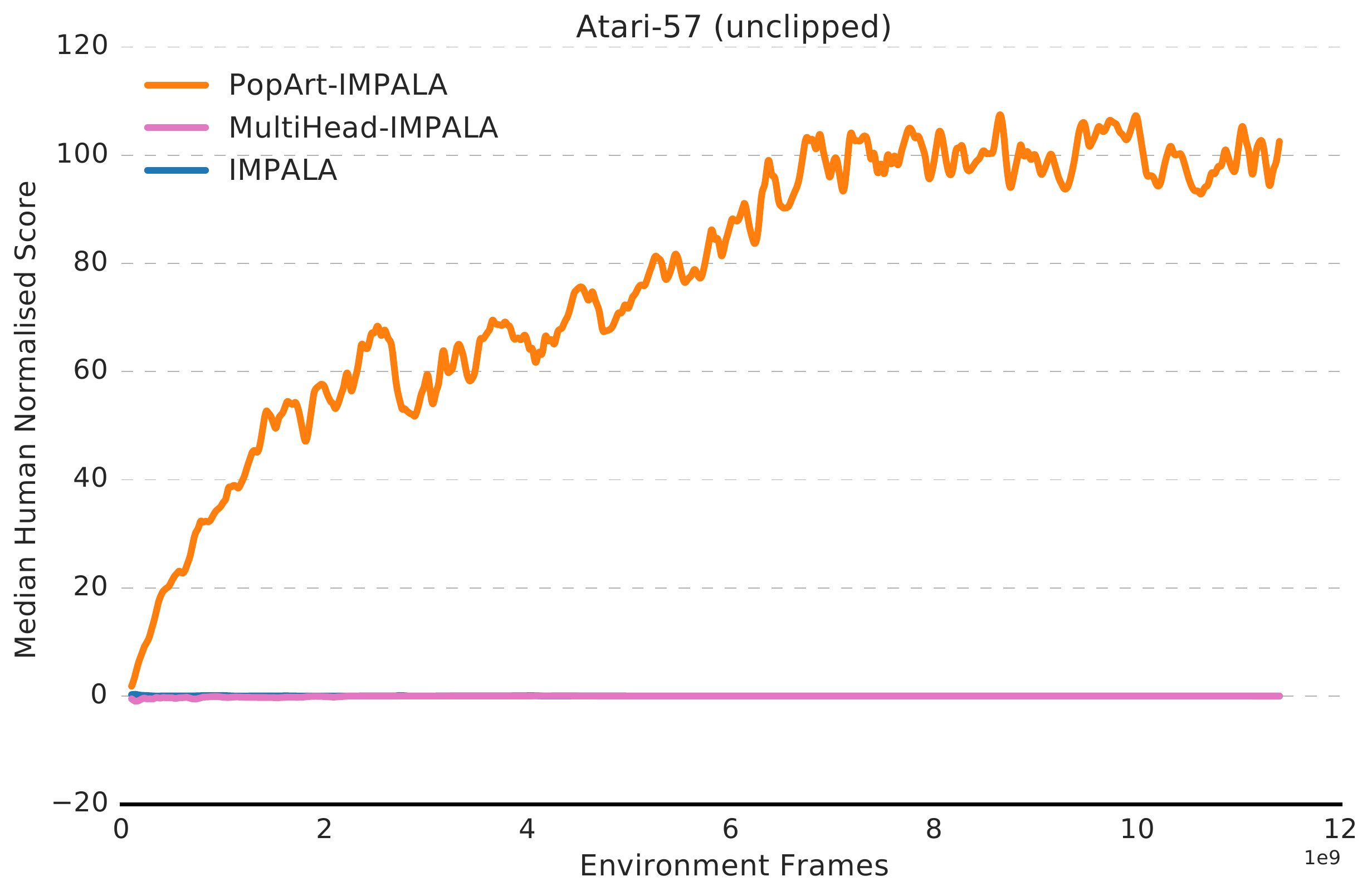}
\caption{\textbf{Atari-57 (unclipped)}: Median human normalised score across all Atari levels, as a function of the total number of frames seen by the agents across all levels. We here compare the same set of agents as in Figure 1, but now all agents are trained without using reward clipping. The approximately flat lines corresponding to the baselines mean no learning at all on at least 50\% of the games.}
\label{fig:atari-curve-unclipped}
\end{figure}

\subsection{Normalisation statistics}

\begin{figure}
\centering
\includegraphics[width=\linewidth]{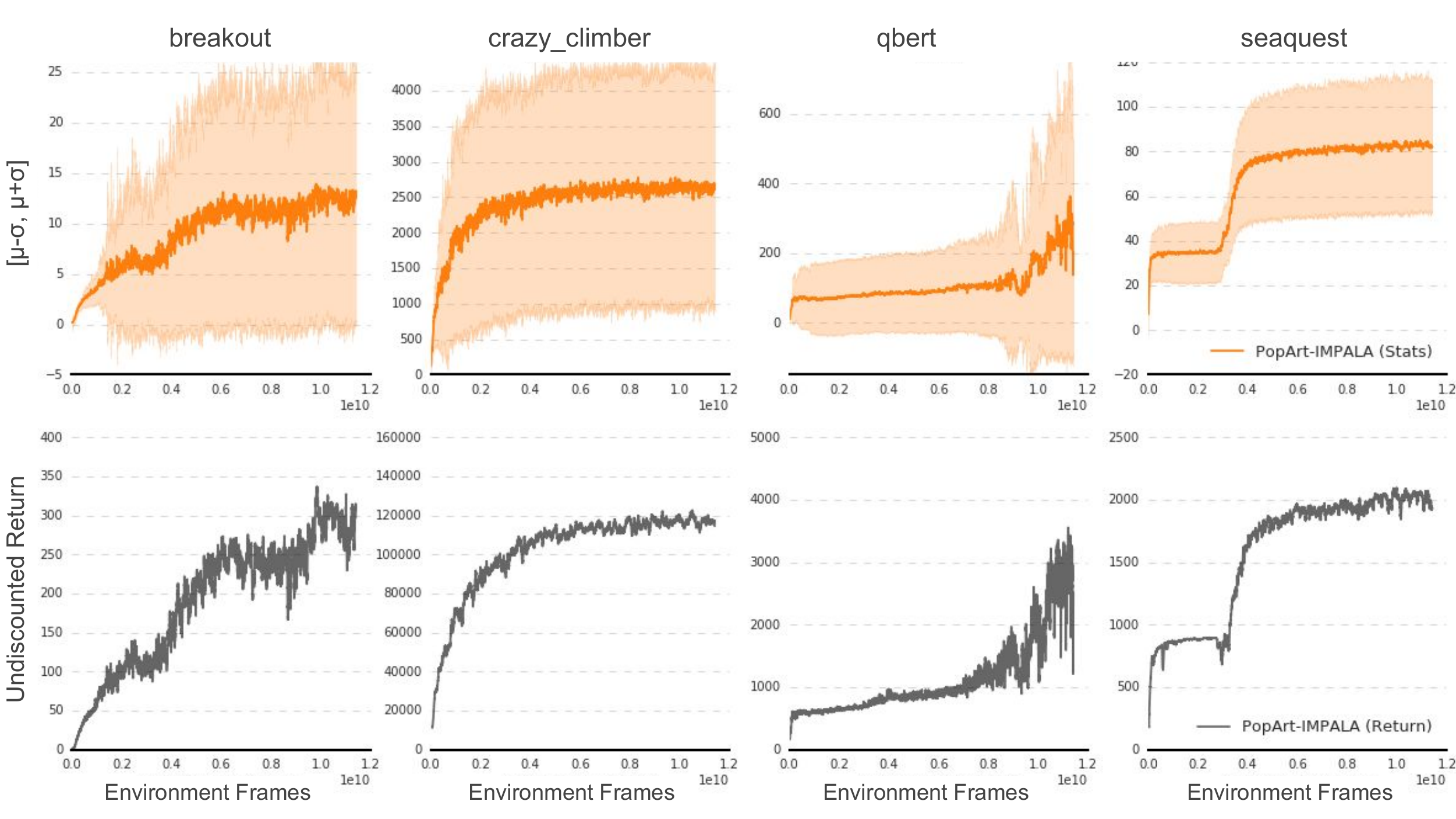}
\caption{\textbf{Normalisation statistics}: Top: learned statistics, without reward clipping, for four distinct Atari games. The shaded region is $[\mu - \sigma, \mu + \sigma]$. Bottom: undiscounted returns.}
\label{fig:stats}
\end{figure}

It is insightful to observe the different normalisation statistics across games, and how they adapt during training. Figure \ref{fig:stats} (top row) plots the \textit{shift} $\mu$ for a selection of Atari games, in the unclipped training regime. The \textit{scale} $\sigma$ is visualised in the same figure by shading the area in the range $[\mu - \sigma, \mu + \sigma]$. The statistics  differ by orders of magnitude across games: in \texttt{crazy\_climber} the shift exceeds 2500, while in \texttt{bowling} it never goes above 15. The adaptivity of the proposed normalisation emerges clearly in \texttt{crazy\_climber} and \texttt{qbert}, where the statistics span multiple orders of magnitude during training. The bottom row in Figure \ref{fig:stats} shows the corresponding agent's undiscounted episode return: it follows the same patterns as the statistics (with differences in magnitude due to discounting). Finally note how the statistics can even track the instabilities in the agent's performance, as in \texttt{qbert}. 

\subsection{DmLab-30 results}

\begin{figure}[t]
\centering
\includegraphics[width=\linewidth]{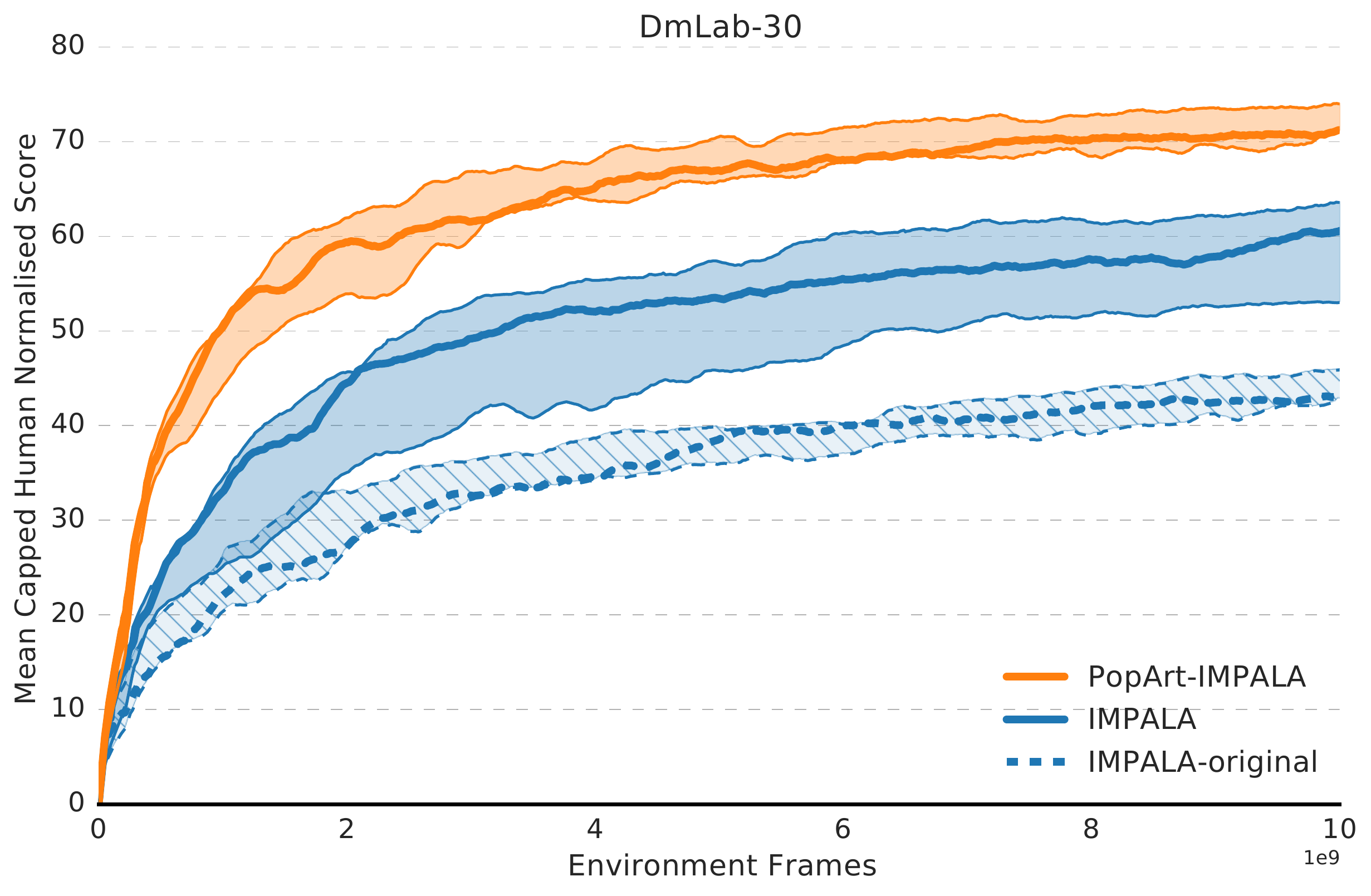}
\caption{\textbf{DmLab-30}. Mean capped human normalised score of IMPALA (blue) and PopArt-IMPALA (orange), across the DmLab-30 benchmark as function of the number of frames (summed across all levels). Shaded region is bounded by best and worse run among 3 PBT experiments. For reference, we also plot the performance of IMPALA with the limited action set from the original paper (dashed).}
\label{fig:dmlab-curve}
\end{figure}

Figure \ref{fig:dmlab-curve} shows, as a function of the total number of frames processed by each agent, the mean human normalised performance across all 30 DeepMind Lab levels, where each level's score is capped at 100\% . For all agents, we ran three independent PBT experiments. In Figure \ref{fig:dmlab-curve} we plot the learning curves for each experiment and, for each agent, fill in the area between best and worse experiment.

Compared to the original paper, our IMPALA baseline uses a richer action set, that includes more possible horizontal rotations, and vertical rotations (details in Appendix).  Fine-grained horizontal control is useful on \texttt{lasertag} levels, while vertical rotations are \textit{necessary} for a few \texttt{psychlab} levels. Note that this new baseline (solid blue in Figure \ref{fig:dmlab-curve}) performs much better than the original IMPALA agent, which we also train and report for completeness (dashed blue).  Including PopArt normalisation (in orange) on top of our baseline results in largely improved scores. Note how agents achieve clearly separated performance levels, with the new action set dominating the original paper's one, and with PopArt-IMPALA dominating IMPALA for all three replications of the experiment. 

\begin{figure}[t]
\centering
\includegraphics[width=\linewidth]{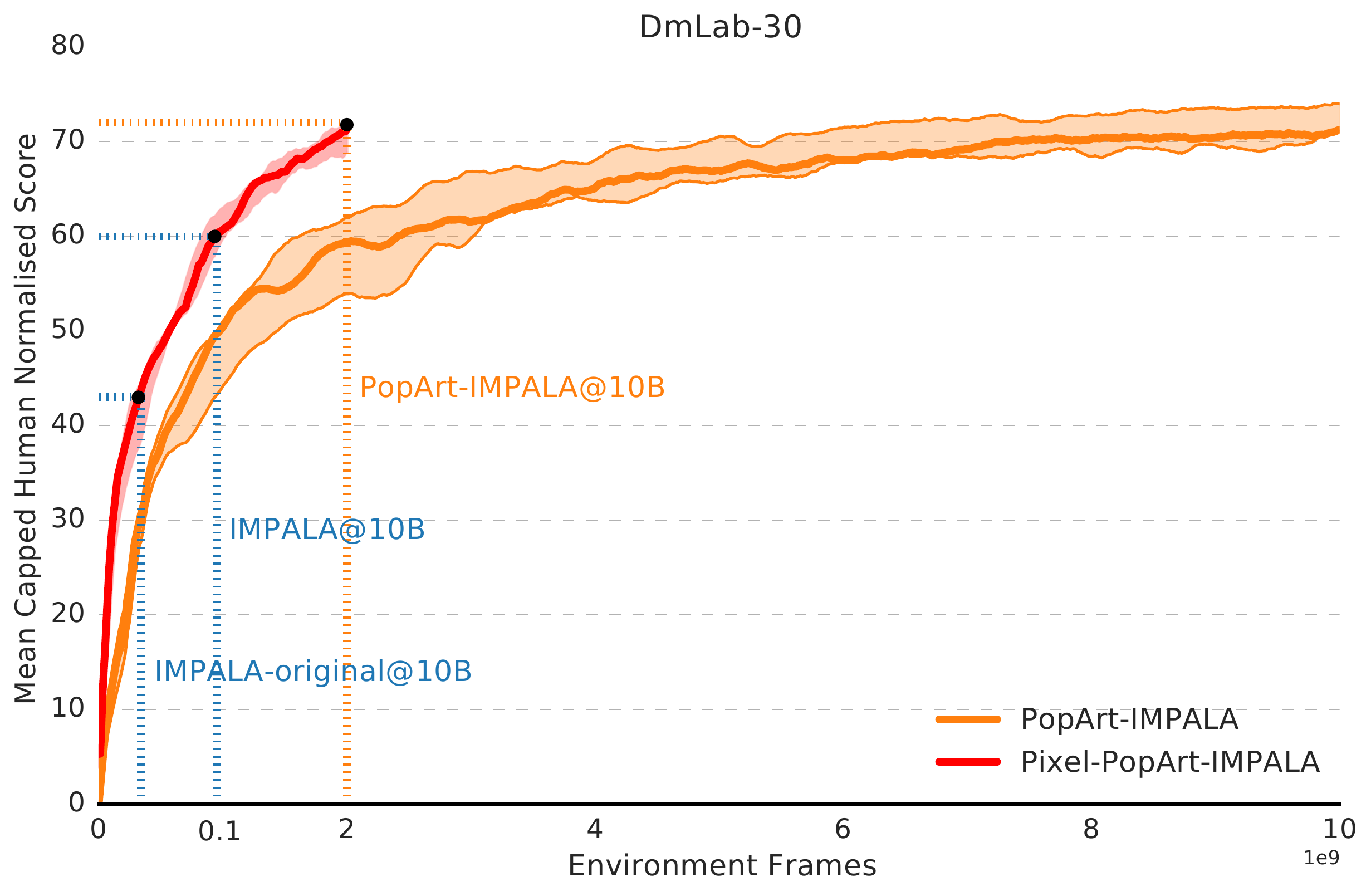}
\caption{\textbf{DmLab-30 (with pixel control)}. Mean capped human normalised score of PopArt-IMPALA with pixel control (red), across the DmLab-30 benchmark as function of the total number of frames across all tasks. Shaded region is bounded by best and worse run among 3 PBT experiments. Dotted lines mark the point where Pixel-PopArt-IMPALA matches PopArt-IMPALA and the two IMPALA baselines.}
\label{fig:dmlab-pixel-main}
\end{figure}

\subsection{Extensions}

In this section, we explore the combination of the proposed PopArt-IMPALA agent with pixel control~\cite{UNREAL}, to further improve data efficiency, and make training IMPALA-like agents on large multi-task benchmarks cheaper and more practical. Pixel control is an unsupervised auxiliary task introduced to help learning good state representations. As shown in Figure \ref{fig:dmlab-pixel-main}, the combination of PopArt-IMPALA with pixel control (red line) allows to match the final performance of the vanilla PopArt-IMPALA (orange line) with a fraction of the data ($\sim2B$ frames). This is on top of the large improvement in data efficiency already provided by PopArt, meaning that the pixel control augmented PopArt-IMPALA needs less than $1/10$-th of the data to match our own IMPALA baseline's performance (and $1/30$-th of the frames to match the original published IMPALA). Importantly, since both PopArt and Pixel Control only add a very small computational cost, this improvement in data efficiency directly translates in a large reduction of the cost of training IMPALA agents on large multi-task benchmarks. Note, finally, that other orthogonal advances in deep RL could also be combined to further improve performance, similarly to what was done by Rainbow \cite{Rainbow}, in the context of value-based reinforcement learning.

\subsection{Implementation notes}

We implemented all agents in TensorFlow. For each batch of rollouts processed by the learner, we average the $G_t^v$ targets within a rollout, and for each rollout in the batch we perform one online update of PopArt's normalisation statistics with \textit{decay} $\beta=3 \times 10^{-4}$. Note that $\beta$ didn't require any tuning. To prevent numerical issues, we clip the \textit{scale} $\sigma$ in the range $[0.0001, 1e6]$. We do not back-propagate gradients into $\mu$ and $\sigma$, exclusively updated as in Equation \ref{stats-rule}. The weights $\mathbf{W}$ of the last layer of the value function are updated according to Equation \ref{popupdate} and \ref{popart-updates-multi}. Note that we first apply the actor-critic updates (\ref{popart-updates-multi}), then update the statistics (\ref{stats-rule}), finally apply output preserving updates (\ref{popupdate}). For more \textit{just-in-time} rescaling of updates we can invert this order, but this wasn't necessary. As anticipated, in all experiments we used population-based training (PBT) to adapt hyperparameters during training \cite{Jaderberg:2017}. As in the IMPALA paper, we use PBT to tune \texttt{learning\_rate}, \texttt{entropy\_cost}, the optimiser's \texttt{epsilon}, and---in the Atari experiments---the \texttt{max\_gradient\_norm}. In Atari-57 we used populations of 24 instances, in DmLab-30 just 8 instances.  All hyperparameters are reported in the Appendix.

\section{Discussion}

In this paper we propose a scale-invariant actor-critic algorithm that enables significantly improved performance in multi-task reinforcement learning settings. Being able to acquire knowledge about a wide range of facts and skills has been long considered an essential feature for an RL agent to demonstrate intelligent behaviour \cite{Horde,Degris:2012,Legg}. To ask our algorithms to master multiple tasks is therefore a natural step as we progress towards increasingly powerful agents. 

The wide-spread adoption of deep learning in RL is quite timely in this regard, since sharing parts of a neural network across multiple tasks is also a powerful way of building robust representations. This is particularly important for RL, because rewards on individual tasks can be sparse, and therefore sharing representations across tasks can be vital to bootstrap learning. Several agents~\cite{UNREAL,AuxTasksFPS,LossItsOwnReward,AuxTasksDepth}
demonstrated this by improving performance on a single external task by learning off-policy about auxiliary tasks defined on the same stream of experience (e.g. pixel control, immediate reward prediction or auto-encoding).

Multi-task learning, as considered in this paper, where we get to execute, in parallel, the policies learned for each task, has potential additional benefits, including deep exploration \cite{Osband:2016}, and policy composition \cite{Mankowitz:2018,ComposeOpt}. By learning on-policy about tasks, it may also be easier to scale to much more diverse tasks: if we only learn about some task off-policy from experience generated pursuing a very different one, we might never observe any reward. A limitation of our approach is that it can be complicated to implement parallel learning outside of simulation, but recent work on parallel training of robots \cite{Levine:2016} suggests that this is not necessarily an insurmountable obstacle if sufficient resources are available.

Adoption of parallel multi-task RL has up to now been fairly limited. That the scaling issues considered in this paper, may have been a factor in the limited adoption is indicated by the wider use of this kind of learning in supervised settings \cite{MultiLangTranslation,MultiSL,CrossStitch,JointManyTask}, where loss functions are naturally well scaled (e.g. cross entropy), or can be easily scaled thanks to the stationarity of the training distribution.  We therefore hope and believe that the work presented here can enable more research on multi-task RL. 

We also believe that PopArt's adaptive normalisation can be combined with other research in multi-task reinforcement learning, that previously did not scale as effectively to large numbers of diverse tasks. We highlight as potential candidates policy distillation \cite{ActorMimic,Distillation,Kickstarting,Distral} and active sampling of the task distribution the agent trains on \cite{SharmaMultiTask}. The combination of PopArt-IMPALA with active sampling might be particularly promising since it may allow a more efficient use of the parallel data generation, by focusing it on the task most amenable for learning. Elastic weight consolidation \cite{EWC} and other work from the continual learning literature \cite{RingContinualPhD94,mcclelland95} might also be adapted to parallel learning setups to reduce interference \cite{Catastrophic} among tasks.

\begin{small}

\bibliographystyle{aaai}
\bibliography{arxiv}

\begin{thebibliography}{}

\bibitem[\protect\citeauthoryear{Bacon, Harb, and Precup}{2017}]{OptionCritic}
Bacon, P.; Harb, J.; and Precup, D.
\newblock 2017.
\newblock The option-critic architecture.
\newblock {\em AAAI Conference on Artificial Intelligence}.

\bibitem[\protect\citeauthoryear{Beattie \bgroup et al\mbox.\egroup
  }{2016}]{DmLab}
Beattie, C.; Leibo, J.~Z.; Teplyashin, D.; Ward, T.; Wainwright, M.;
  K{\"{u}}ttler, H.; Lefrancq, A.; Green, S.; Vald{\'{e}}s, V.; Sadik, A.;
  Schrittwieser, J.; Anderson, K.; York, S.; Cant, M.; Cain, A.; Bolton, A.;
  Gaffney, S.; King, H.; Hassabis, D.; Legg, S.; and Petersen, S.
\newblock 2016.
\newblock Deepmind lab.
\newblock {\em CoRR} abs/1612.03801.

\bibitem[\protect\citeauthoryear{Bellemare \bgroup et al\mbox.\egroup
  }{2013}]{bellemare2013arcade}
Bellemare, M.~G.; Naddaf, Y.; Veness, J.; and Bowling, M.
\newblock 2013.
\newblock The arcade learning environment: An evaluation platform for general
  agents.
\newblock {\em JAIR}.

\bibitem[\protect\citeauthoryear{Bellman}{1957}]{bellmanMDP}
Bellman, R.
\newblock 1957.
\newblock A markovian decision process.
\newblock {\em Journal of Mathematics and Mechanics}.

\bibitem[\protect\citeauthoryear{Brunskill and
  Li}{2013}]{Brunskill2013SampleCO}
Brunskill, E., and Li, L.
\newblock 2013.
\newblock Sample complexity of multi-task reinforcement learning.
\newblock {\em CoRR} abs/1309.6821.

\bibitem[\protect\citeauthoryear{Caruana}{1998}]{caruana1998multitask}
Caruana, R.
\newblock 1998.
\newblock Multitask learning.
\newblock In {\em Learning to learn}.

\bibitem[\protect\citeauthoryear{Collobert and Weston}{2008}]{MultiNLP}
Collobert, R., and Weston, J.
\newblock 2008.
\newblock A unified architecture for natural language processing: Deep neural
  networks with multitask learning.
\newblock In {\em ICML}.

\bibitem[\protect\citeauthoryear{Degris and Modayil}{2012}]{Degris:2012}
Degris, T., and Modayil, J.
\newblock 2012.
\newblock Scaling-up knowledge for a cognizant robot.
\newblock In {\em AAAI Spring Symposium: Designing Intelligent Robots}.

\bibitem[\protect\citeauthoryear{Duan \bgroup et al\mbox.\egroup
  }{2016}]{duan2016benchmarking}
Duan, Y.; Chen, X.; Houthooft, R.; Schulman, J.; and Abbeel, P.
\newblock 2016.
\newblock Benchmarking deep reinforcement learning for continuous control.
\newblock In {\em ICML}.

\bibitem[\protect\citeauthoryear{Espeholt \bgroup et al\mbox.\egroup
  }{2018}]{IMPALA}
Espeholt, L.; Soyer, H.; Munos, R.; Simonyan, K.; Mnih, V.; Ward, T.; Doron,
  Y.; Firoiu, V.; Harley, T.; Dunning, I.; Legg, S.; and Kavukcuoglu, K.
\newblock 2018.
\newblock Impala: Scalable distributed deep-rl with importance weighted
  actor-learner architectures.
\newblock In {\em ICML}.

\bibitem[\protect\citeauthoryear{French}{1999}]{Catastrophic}
French, R.~M.
\newblock 1999.
\newblock Catastrophic forgetting in connectionist networks.
\newblock {\em Trends in cognitive sciences}.

\bibitem[\protect\citeauthoryear{Gruslys \bgroup et al\mbox.\egroup
  }{2018}]{Reactor}
Gruslys, A.; Azar, M.~G.; Bellemare, M.~G.; and Munos, R.
\newblock 2018.
\newblock The reactor: A sample-efficient actor-critic architecture.
\newblock {\em ICLR}.

\bibitem[\protect\citeauthoryear{Hashimoto \bgroup et al\mbox.\egroup
  }{2016}]{JointManyTask}
Hashimoto, K.; Xiong, C.; Tsuruoka, Y.; and Socher, R.
\newblock 2016.
\newblock A joint many-task model: Growing a neural network for multiple {NLP}
  tasks.
\newblock {\em CoRR} abs/1611.01587.

\bibitem[\protect\citeauthoryear{He \bgroup et al\mbox.\egroup
  }{2015}]{He:2015}
He, K.; Zhang, X.; Ren, S.; and Sun, J.
\newblock 2015.
\newblock Deep residual learning for image recognition.
\newblock {\em arXiv preprint arXiv:1512.03385}.

\bibitem[\protect\citeauthoryear{Hessel \bgroup et al\mbox.\egroup
  }{2018}]{Rainbow}
Hessel, M.; Modayil, J.; van Hasselt, H.; Schaul, T.; Ostrovski, G.; Dabney,
  W.; Horgan, D.; Piot, B.; Azar, M.~G.; and Silver, D.
\newblock 2018.
\newblock Rainbow: Combining improvements in deep reinforcement learning.
\newblock {\em AAAI Conference on Artificial Intelligence}.

\bibitem[\protect\citeauthoryear{Hochreiter and
  Schmidhuber}{1997}]{Hochreiter:1997}
Hochreiter, S., and Schmidhuber, J.
\newblock 1997.
\newblock Long short-term memory.
\newblock {\em Neural computation}.

\bibitem[\protect\citeauthoryear{Jaderberg \bgroup et al\mbox.\egroup
  }{2016}]{UNREAL}
Jaderberg, M.; Mnih, V.; Czarnecki, W.~M.; Schaul, T.; Leibo, J.~Z.; Silver,
  D.; and Kavukcuoglu, K.
\newblock 2016.
\newblock Reinforcement learning with unsupervised auxiliary tasks.
\newblock {\em CoRR} abs/1611.05397.

\bibitem[\protect\citeauthoryear{Jaderberg \bgroup et al\mbox.\egroup
  }{2017}]{Jaderberg:2017}
Jaderberg, M.; Dalibard, V.; Osindero, S.; Czarnecki, W.~M.; Donahue, J.;
  Razavi, A.; Vinyals, O.; Green, T.; Dunning, I.; Simonyan, K.; Fernando, C.;
  and Kavukcuoglu, K.
\newblock 2017.
\newblock Population based training of neural networks.
\newblock {\em CoRR} abs/1711.09846.

\bibitem[\protect\citeauthoryear{Johnson \bgroup et al\mbox.\egroup
  }{2017}]{MultiLangTranslation}
Johnson, M.; Schuster, M.; Le, Q.~V.; Krikun, M.; Wu, Y.; Chen, Z.; Thorat, N.;
  Vi{\'{e}}gas, F.~B.; Wattenberg, M.; Corrado, G.; Hughes, M.; and Dean, J.
\newblock 2017.
\newblock Google's multilingual neural machine translation system: Enabling
  zero-shot translation.
\newblock {\em Transactions of the Association for Computational Linguistics}
  5.

\bibitem[\protect\citeauthoryear{Kirkpatrick \bgroup et al\mbox.\egroup
  }{2017}]{EWC}
Kirkpatrick, J.; Pascanu, R.; Rabinowitz, N.; Veness, J.; Desjardins, G.; Rusu,
  A.~A.; Milan, K.; Quan, J.; Ramalho, T.; Grabska-Barwinska, A.; Hassabis, D.;
  Clopath, C.; Kumaran, D.; and Hadsell, R.
\newblock 2017.
\newblock Overcoming catastrophic forgetting in neural networks.
\newblock {\em PNAS}.

\bibitem[\protect\citeauthoryear{Lample and Chaplot}{2016}]{AuxTasksFPS}
Lample, G., and Chaplot, D.~S.
\newblock 2016.
\newblock Playing {FPS} games with deep reinforcement learning.
\newblock {\em CoRR} abs/1609.05521.

\bibitem[\protect\citeauthoryear{Legg and Hutter}{2007}]{Legg}
Legg, S., and Hutter, M.
\newblock 2007.
\newblock Universal intelligence: A definition of machine intelligence.
\newblock {\em Minds Mach.}

\bibitem[\protect\citeauthoryear{Levine \bgroup et al\mbox.\egroup
  }{2016}]{Levine:2016}
Levine, S.; Pastor, P.; Krizhevsky, A.; and Quillen, D.
\newblock 2016.
\newblock Learning hand-eye coordination for robotic grasping with large-scale
  data collection.
\newblock In {\em ISER}.

\bibitem[\protect\citeauthoryear{Lillicrap \bgroup et al\mbox.\egroup
  }{2016}]{Lillicrap2016}
Lillicrap, T.; Hunt, J.; Pritzel, A.; Heess, N.; Erez, T.; Tassa, Y.; Silver,
  D.; and Wierstra, D.
\newblock 2016.
\newblock Continuous control with deep reinforcement learning.
\newblock In {\em ICLR}.

\bibitem[\protect\citeauthoryear{Lu \bgroup et al\mbox.\egroup
  }{2016}]{MultiSL}
Lu, Y.; Kumar, A.; Zhai, S.; Cheng, Y.; Javidi, T.; and Feris, R.~S.
\newblock 2016.
\newblock Fully-adaptive feature sharing in multi-task networks with
  applications in person attribute classification.
\newblock {\em CoRR} abs/1611.05377.

\bibitem[\protect\citeauthoryear{Mahmood}{2017}]{Mahmood:2017}
Mahmood, A.
\newblock 2017.
\newblock Incremental off-policy reinforcement learning algorithms.
\newblock {\em Ph.D. UAlberta}.

\bibitem[\protect\citeauthoryear{Mankowitz \bgroup et al\mbox.\egroup
  }{2018}]{Mankowitz:2018}
Mankowitz, D.~J.; Z{\'{\i}}dek, A.; Barreto, A.; Horgan, D.; Hessel, M.; Quan,
  J.; Oh, J.; van Hasselt, H.; Silver, D.; and Schaul, T.
\newblock 2018.
\newblock Unicorn: Continual learning with a universal, off-policy agent.
\newblock {\em CoRR} abs/1802.08294.

\bibitem[\protect\citeauthoryear{Mcclelland, Mcnaughton, and
  O'Reilly}{1995}]{mcclelland95}
Mcclelland, J.~L.; Mcnaughton, B.~L.; and O'Reilly, R.~C.
\newblock 1995.
\newblock Why there are complementary learning systems in the hippocampus and
  neocortex: {I}nsights from the successes and failures of connectionist models
  of learning and memory.
\newblock {\em Psychological Review}.

\bibitem[\protect\citeauthoryear{Mirowski \bgroup et al\mbox.\egroup
  }{2016}]{AuxTasksDepth}
Mirowski, P.; Pascanu, R.; Viola, F.; Soyer, H.; Ballard, A.~J.; Banino, A.;
  Denil, M.; Goroshin, R.; Sifre, L.; Kavukcuoglu, K.; Kumaran, D.; and
  Hadsell, R.
\newblock 2016.
\newblock Learning to navigate in complex environments.
\newblock {\em CoRR} abs/1611.03673.

\bibitem[\protect\citeauthoryear{Misra \bgroup et al\mbox.\egroup
  }{2016}]{CrossStitch}
Misra, I.; Shrivastava, A.; Gupta, A.; and Hebert, M.
\newblock 2016.
\newblock Cross-stitch networks for multi-task learning.
\newblock {\em CoRR} abs/1604.03539.

\bibitem[\protect\citeauthoryear{Mnih \bgroup et al\mbox.\egroup
  }{2015}]{Mnih2015}
Mnih, V.; Kavukcuoglu, K.; Silver, D.; Rusu, A.~A.; Veness, J.; Bellemare,
  M.~G.; Graves, A.; Riedmiller, M.; Fidjeland, A.~K.; Ostrovski, G.; Petersen,
  S.; Beattie, C.; Sadik, A.; Antonoglou, I.; King, H.; Kumaran, D.; Wierstra,
  D.; Legg, S.; and Hassabis, D.
\newblock 2015.
\newblock Human-level control through deep reinforcement learning.
\newblock {\em Nature}.

\bibitem[\protect\citeauthoryear{Mnih \bgroup et al\mbox.\egroup
  }{2016}]{Mnih2016}
Mnih, V.; Badia, A.~P.; Mirza, M.; Graves, A.; Lillicrap, T.; Harley, T.;
  Silver, D.; and Kavukcuoglu, K.
\newblock 2016.
\newblock Asynchronous methods for deep reinforcement learning.
\newblock In {\em ICML}.

\bibitem[\protect\citeauthoryear{Nair \bgroup et al\mbox.\egroup
  }{2015}]{Nair2015}
Nair, A.; Srinivasan, P.; Blackwell, S.; Alcicek, C.; Fearon, R.; De~Maria, A.;
  Panneershelvam, V.; Suleyman, M.; Beattie, C.; Petersen, S.; Legg, S.; Mnih,
  V.; Kavukcuoglu, K.; and Silver, D.
\newblock 2015.
\newblock Massively parallel methods for deep reinforcement learning.
\newblock {\em arXiv preprint arXiv:1507.04296}.

\bibitem[\protect\citeauthoryear{Osband \bgroup et al\mbox.\egroup
  }{2016}]{Osband:2016}
Osband, I.; Blundell, C.; Pritzel, A.; and Van~Roy, B.
\newblock 2016.
\newblock Deep exploration via bootstrapped {DQN}.
\newblock In {\em NIPS}.

\bibitem[\protect\citeauthoryear{Parisotto, Ba, and
  Salakhutdinov}{2015}]{ActorMimic}
Parisotto, E.; Ba, L.~J.; and Salakhutdinov, R.
\newblock 2015.
\newblock Actor-mimic: Deep multitask and transfer reinforcement learning.
\newblock {\em CoRR} abs/1511.06342.

\bibitem[\protect\citeauthoryear{Precup, Sutton, and Singh}{2000}]{Precup:2000}
Precup, D.; Sutton, R.~S.; and Singh, S.~P.
\newblock 2000.
\newblock Eligibility traces for off-policy policy evaluation.
\newblock In {\em ICML}.

\bibitem[\protect\citeauthoryear{Ring}{1994}]{RingContinualPhD94}
Ring, M.
\newblock 1994.
\newblock Continual learning in reinforcement environments.

\bibitem[\protect\citeauthoryear{Romera-Paredes \bgroup et al\mbox.\egroup
  }{2013}]{MultiPontil}
Romera-Paredes, B.; Aung, H.; Bianchi-Berthouze, N.; and Pontil, M.
\newblock 2013.
\newblock Multilinear multitask learning.
\newblock In {\em ICML}.

\bibitem[\protect\citeauthoryear{Rusu \bgroup et al\mbox.\egroup
  }{2015}]{Distillation}
Rusu, A.~A.; Colmenarejo, S.~G.; G{\"{u}}l{\c{c}}ehre, {\c{C}}.; Desjardins,
  G.; Kirkpatrick, J.; Pascanu, R.; Mnih, V.; Kavukcuoglu, K.; and Hadsell, R.
\newblock 2015.
\newblock Policy distillation.
\newblock {\em CoRR} abs/1511.06295.

\bibitem[\protect\citeauthoryear{Rusu \bgroup et al\mbox.\egroup
  }{2016}]{Progressive}
Rusu, A.~A.; Rabinowitz, N.~C.; Desjardins, G.; Soyer, H.; Kirkpatrick, J.;
  Kavukcuoglu, K.; Pascanu, R.; and Hadsell, R.
\newblock 2016.
\newblock Progressive neural networks.
\newblock {\em CoRR} abs/1606.04671.

\bibitem[\protect\citeauthoryear{Schmidhuber}{1990}]{Schmidhuber90}
Schmidhuber, J.
\newblock 1990.
\newblock An on-line algorithm for dynamic reinforcement learning and planning
  in reactive environments.
\newblock In {\em IJCNN}.

\bibitem[\protect\citeauthoryear{Schmitt \bgroup et al\mbox.\egroup
  }{2018}]{Kickstarting}
Schmitt, S.; Hudson, J.~J.; Z{\'{\i}}dek, A.; Osindero, S.; Doersch, C.;
  Czarnecki, W.~M.; Leibo, J.~Z.; K{\"{u}}ttler, H.; Zisserman, A.; Simonyan,
  K.; and Eslami, S. M.~A.
\newblock 2018.
\newblock Kickstarting deep reinforcement learning.
\newblock {\em CoRR} abs/1803.03835.

\bibitem[\protect\citeauthoryear{Schulman \bgroup et al\mbox.\egroup
  }{2015}]{SchulmanTRPO}
Schulman, J.; Levine, S.; Moritz, P.; Jordan, M.~I.; and Abbeel, P.
\newblock 2015.
\newblock Trust region policy optimization.
\newblock {\em CoRR} abs/1502.05477.

\bibitem[\protect\citeauthoryear{Schulman \bgroup et al\mbox.\egroup
  }{2017}]{SchulmanPPO}
Schulman, J.; Wolski, F.; Dhariwal, P.; Radford, A.; and Klimov, O.
\newblock 2017.
\newblock Proximal policy optimization algorithms.
\newblock {\em CoRR} abs/1707.06347.

\bibitem[\protect\citeauthoryear{Sharma and Ravindran}{2017}]{SharmaMultiTask}
Sharma, S., and Ravindran, B.
\newblock 2017.
\newblock Online multi-task learning using active sampling.
\newblock {\em CoRR} abs/1702.06053.

\bibitem[\protect\citeauthoryear{Shelhamer \bgroup et al\mbox.\egroup
  }{2016}]{LossItsOwnReward}
Shelhamer, E.; Mahmoudieh, P.; Argus, M.; and Darrell, T.
\newblock 2016.
\newblock Loss is its own reward: Self-supervision for reinforcement learning.
\newblock {\em CoRR} abs/1612.07307.

\bibitem[\protect\citeauthoryear{Silver \bgroup et al\mbox.\egroup
  }{2016}]{Silver_2016}
Silver, D.; Huang, A.; Maddison, C.~J.; Guez, A.; Sifre, L.; van~den Driessche,
  G.; Schrittwieser, J.; Antonoglou, I.; Panneershelvam, V.; Lanctot, M.;
  Dieleman, S.; Grewe, D.; Nham, J.; Kalchbrenner, N.; Sutskever, I.;
  Lillicrap, T.; Leach, M.; Kavukcuoglu, K.; Graepel, T.; and Hassabis, D.
\newblock 2016.
\newblock Mastering the game of {Go} with deep neural networks and tree search.
\newblock {\em Nature}.

\bibitem[\protect\citeauthoryear{Silver \bgroup et al\mbox.\egroup
  }{2017}]{AlphaZero}
Silver, D.; Hubert, T.; Schrittwieser, J.; Antonoglou, I.; Lai, M.; Guez, A.;
  Lanctot, M.; Sifre, L.; Kumaran, D.; Graepel, T.; Lillicrap, T.~P.; Simonyan,
  K.; and Hassabis, D.
\newblock 2017.
\newblock Mastering chess and shogi by self-play with a general reinforcement
  learning algorithm.
\newblock {\em CoRR} abs/1712.01815.

\bibitem[\protect\citeauthoryear{Sutton and Barto}{2018}]{SB2018}
Sutton, R.~S., and Barto, A.~G.
\newblock 2018.
\newblock {\em Reinforcement Learning: An Introduction}.
\newblock MIT press.

\bibitem[\protect\citeauthoryear{Sutton \bgroup et al\mbox.\egroup
  }{2011}]{Horde}
Sutton, R.~S.; Modayil, J.; Delp, M.; Degris, T.; Pilarski, P.~M.; White, A.;
  and Precup, D.
\newblock 2011.
\newblock Horde: A scalable real-time architecture for learning knowledge from
  unsupervised sensorimotor interaction.
\newblock In {\em AAMAS}.

\bibitem[\protect\citeauthoryear{Sutton \bgroup et al\mbox.\egroup
  }{2014}]{Sutton:2014}
Sutton, R.~S.; Mahmood, A.~R.; Precup, D.; and van Hasselt, H.
\newblock 2014.
\newblock A new q($\lambda$) with interim forward view and {Monte Carlo}
  equivalence.
\newblock In {\em ICML}.

\bibitem[\protect\citeauthoryear{Teh \bgroup et al\mbox.\egroup
  }{2017}]{Distral}
Teh, Y.~W.; Bapst, V.; Czarnecki, W.~M.; Quan, J.; Kirkpatrick, J.; Hadsell,
  R.; Heess, N.; and Pascanu, R.
\newblock 2017.
\newblock Distral: Robust multitask reinforcement learning.
\newblock {\em CoRR} abs/1707.04175.

\bibitem[\protect\citeauthoryear{Thrun}{1996}]{thrun1996learning}
Thrun, S.
\newblock 1996.
\newblock Is learning the n-th thing any easier than learning the first?
\newblock In {\em NIPS}.

\bibitem[\protect\citeauthoryear{Thrun}{2012}]{thrun2012explanation}
Thrun, S.
\newblock 2012.
\newblock {\em Explanation-based neural network learning: A lifelong learning
  approach}.
\newblock Springer.

\bibitem[\protect\citeauthoryear{Todorov}{2009}]{ComposeOpt}
Todorov, E.
\newblock 2009.
\newblock Compositionality of optimal control laws.
\newblock In {\em NIPS}.

\bibitem[\protect\citeauthoryear{van Hasselt \bgroup et al\mbox.\egroup
  }{2016}]{PopNIPS}
van Hasselt, H.; Guez, A.; Hessel, M.; Mnih, V.; and Silver, D.
\newblock 2016.
\newblock Learning values across many orders of magnitude.
\newblock In {\em NIPS}.

\bibitem[\protect\citeauthoryear{{van Hasselt}, Guez, and
  Silver}{2016}]{van2016deep}
{van Hasselt}, H.; Guez, A.; and Silver, D.
\newblock 2016.
\newblock Deep reinforcement learning with double {Q}-learning.
\newblock In {\em AAAI Conference on Artificial Intelligence}.

\bibitem[\protect\citeauthoryear{Watkins}{1989}]{Watkins:1989}
Watkins, C. J. C.~H.
\newblock 1989.
\newblock {\em Learning from Delayed Rewards}.
\newblock Ph.D. Dissertation, King's College, Cambridge, England.

\bibitem[\protect\citeauthoryear{Williams}{1992}]{Williams1992}
Williams, R.
\newblock 1992.
\newblock Simple statistical gradient-following algorithms for connectionist
  reinforcement learning.
\newblock {\em Mach. Learning}.

\end{thebibliography}

\end{small}

\newpage

\section{Appendix}

In this Appendix we report additional details about the results presented in the main text, as well as present additional experiments on the DmLab-30 benchmark. We also report the breakdown per level of the scores of IMPALA and PopArt-IMPALA on the Atari-57 and DmLab-30 benchmarks. Finally, we report the hyperparameters used to train the baseline agents as well as PopArt-IMPALA. These hyperparameters are mostly the same as in Espeholt et al., but we report them for completeness and to ease reproducibility. 

\subsection{Hyper-parameter tuning}

In our experiments we used Population-Based Training (PBT) to tune hyper-parameters. In our DmLab-30 experiments, however, we used smaller populations than in the original IMPALA paper. For completeness, we also report here the results of running PopArt-IMPALA and IMPALA with the larger population size used by Espeholt et al. Due to the increased cost of using larger populations, in this case we will only report one PBT tuning experiment per agent, rather than the 3 reported in the main text. 

The learning curves for both IMPALA and PopArt-IMPALA are shown in Figure \ref{fig:dmlab-poplarge}, together with horizontal dashed lines marking average final performance of the agents trained with the smaller population of just 8 instances. The performance of both IMPALA and PopArt-IMPALA agents at the end of training is very similar whether hyperparameters are tuned with 8 or 24 PBT instances, suggesting that the large populations used for hyper-parameter tuning by Espeholt et al. may not be necessary. 

Note, however, that we have observed larger discrepancies between experiments where small and large population size are used for tuning hyper-parameter, when training the less performing IMPALA agent that used a more limited action set, as presented in the original IMPALA paper.

\vspace{10pt}

\begin{figure}[h!]
\setcounter{figure}{5}
\centering
\includegraphics[width=0.8\linewidth]{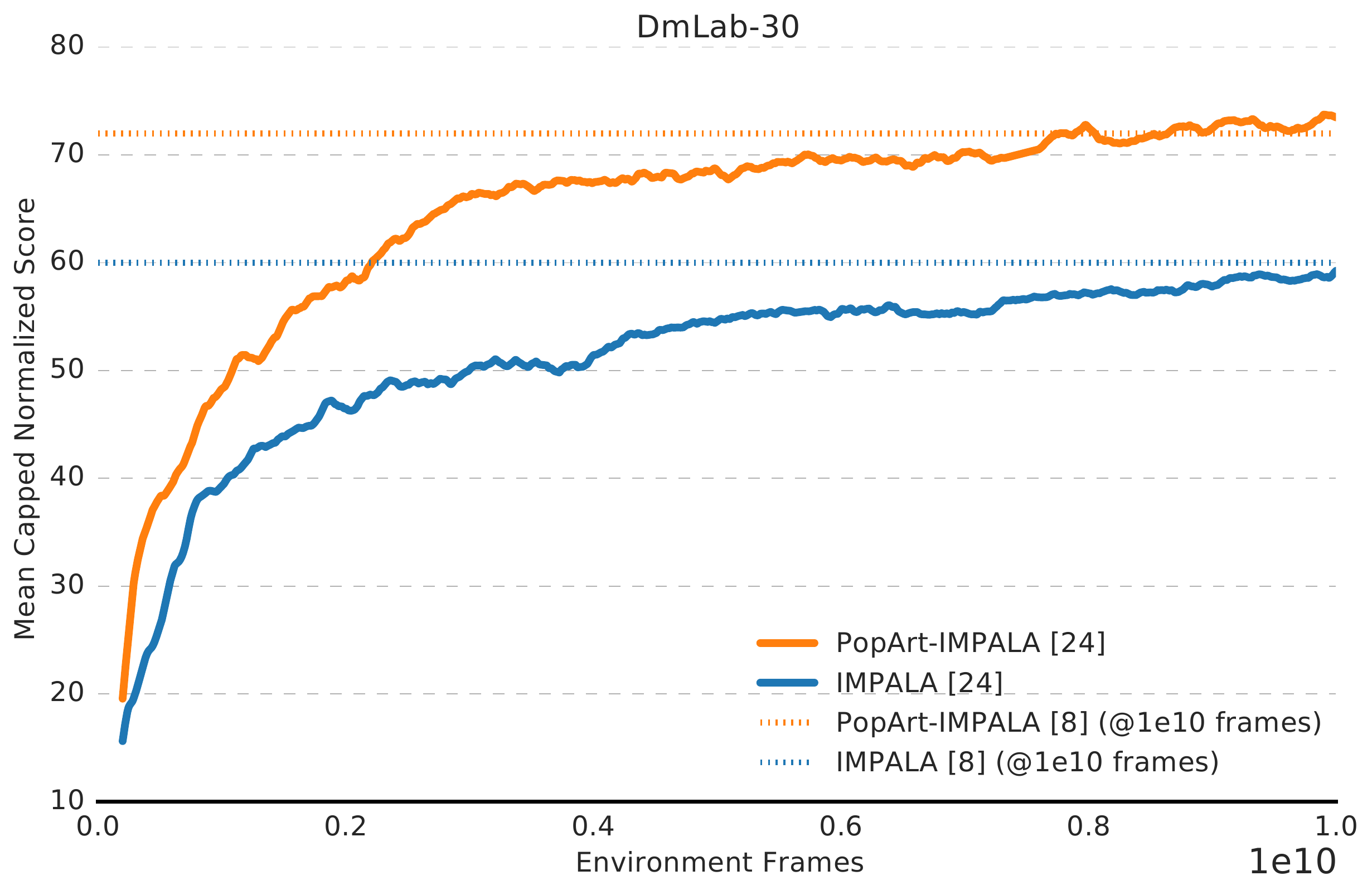}
\caption{\textbf{Larger populations}: mean capped human normalised score across the DmLab-30 benchmark as a function of the total number of frames seen by the agents across all levels. The solid lines plot the performance of IMPALA (blue) and PopArt-IMPALA (orange) when tuning hyperparameters with a large PBT population of 24 instances. Dashed lines correspond to the final performance of these same agents, after 10B frames, in the previous experiments where hyper-parameters were tuned with a population of 8.}
\label{fig:dmlab-poplarge}
\end{figure}

\subsection{Pixel Control}

\begin{figure}[h!]

\centering
\includegraphics[width=0.8\linewidth]{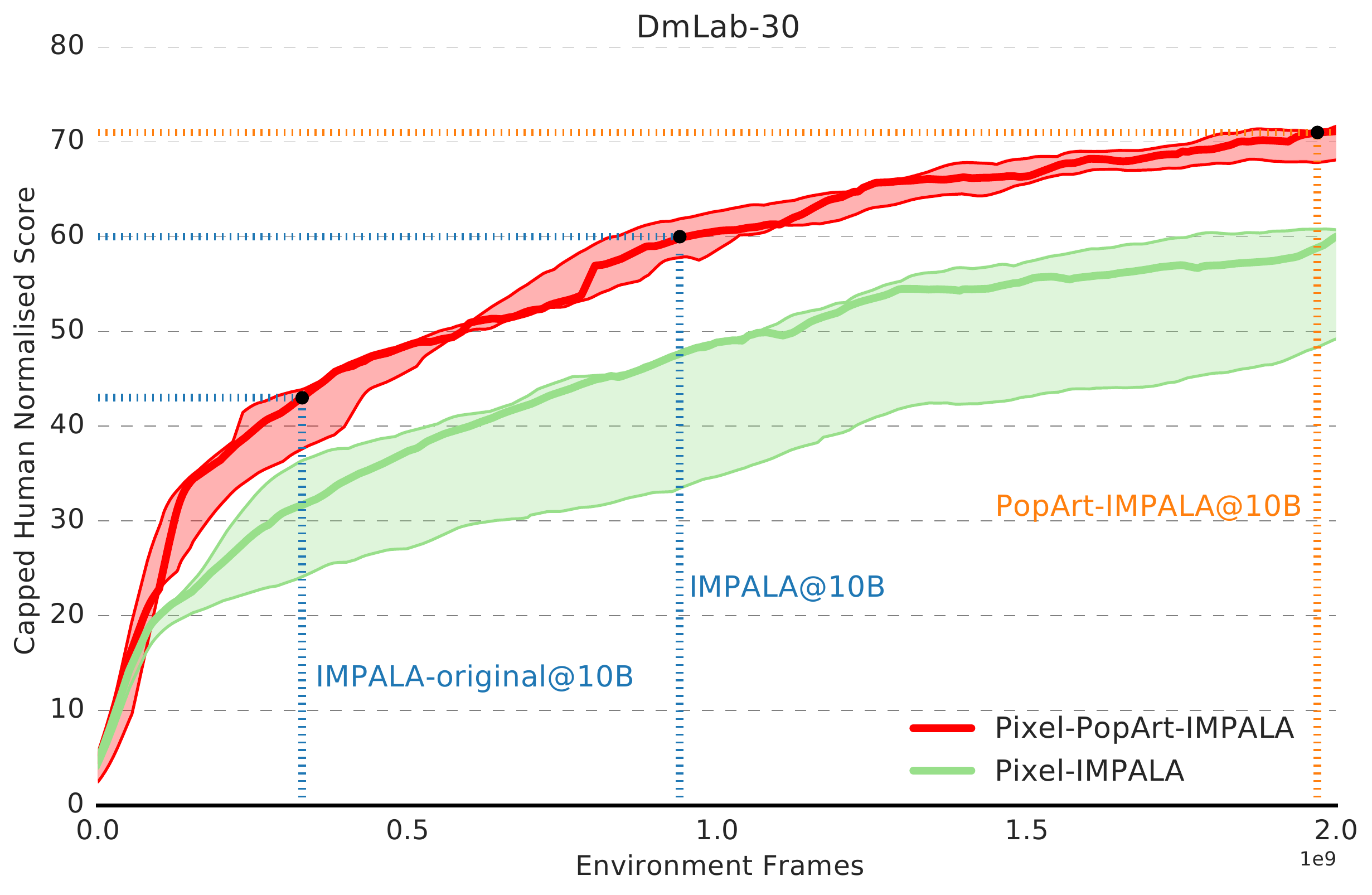}
\caption{\textbf{Pixel Control}: mean capped human normalised score across the DmLab-30 benchmark as a function of the total number of frames (summed across levels). Solid lines plot the performance PopArt-IMPALA (red) and IMPALA (green), after augmenting both with pixel control. Dashed lines mark the point at which Pixel-PopArt-IMPALA matches the final performance of previous agents. Note how, thanks to the improved data efficiency, we train for 2B frames, compared to 10B in previous experiments.}
\label{fig:dmlab-pixel}

\end{figure}

Pixel control \cite{UNREAL} is an unsupervised auxiliary task introduced to help learning good state representations. We report here also the performance of combining Pixel Control with IMPALA without also using PopArt. As shown in Figure \ref{fig:dmlab-pixel}, pixel control increases the performance of both the PopArt-IMPALA agent as well as that of the IMPALA baseline. PopArt still guarantees a noticeable boost in performance, with the median human normalized score of Pixel-PopArt-IMPALA (red line) exceeding the score of Pixel-IMPALA (green line) by approximately 10 points.

We implemented the pixel control task as described in the original paper, only adapting the scheme to the rectangular observations used in DmLab-30. We split the $(72 \times 96)$ observations into a $18 \times 24$ grid of $4 \times 4$ cells. For each location in the grid we define a distinct pseudo-reward $\tilde{r}_{i,j}$, equal to the absolute value of the difference between pixel intensities in consecutive frames, averaged across the 16 pixels of cell $c_{i, j}$. For each cell, we train action values with multi-step Q-learning, accumulating rewards until the end of a rollout and then bootstrapping. We use a discount $\gamma=0.9$.  Learning is fully off-policy on experience generated by the actors, that follow the main policy $\pi$ as usual.

We use a deep deconvolutional network for the action value predictions associated to each pseudo-reward $\tilde{r}_{i,j}$. First, we feed the LSTM's output to a fully connected layer, reshape the output tensor as $6 \times 9 \times 32$, and apply a deconvolution with $3 \times 3$ kernels that outputs a $8 \times 11 \times 32$ tensor. From this, we compute a spatial grid of Q-values using a dueling network architecture: we use a deconvolution with 1 output channel for the state values across the grid and a deconvolution with $|\A|$ channels for the advantage estimates of each cell. Output deconvolutions use $4 \times 4$ kernels with stride $2$. The additional head is only evaluated on the learner, actors do not execute it.

\vspace{10pt}

\subsection{Atari-57 Score breakdown}

In this section we use barplots to report the final performance of the agents on each of the levels in the Atari-57 multi-task benchmark. In order to compute these scores we take the final trained agent and evaluate it with a frozen policy on each of the levels for 200 episodes. The same trained policy is evaluated in all the levels, and the policy is not provided information about the task it's being evaluated on.
For Atari, we compare PopArt-IMPALA, with and without reward clipping, to an IMPALA baseline. In all cases the height of the bars in the plot denote human normalised score. For the Atari results we additionally rescale logarithmically the x-axis, because in this domain games may differ in their normalised performance by several orders of magnitude.

\begin{figure}[h!]
\centering
\caption{\textbf{Atari-57 breakdown}: human normalised score for IMPALA and PopArt-IMPALA, as measured in a separate evaluation phase at the end of training, broken down for the 57 games. For PopArt-IMPALA we report the scores both with and without reward clipping}
\includegraphics[width=0.87\linewidth]{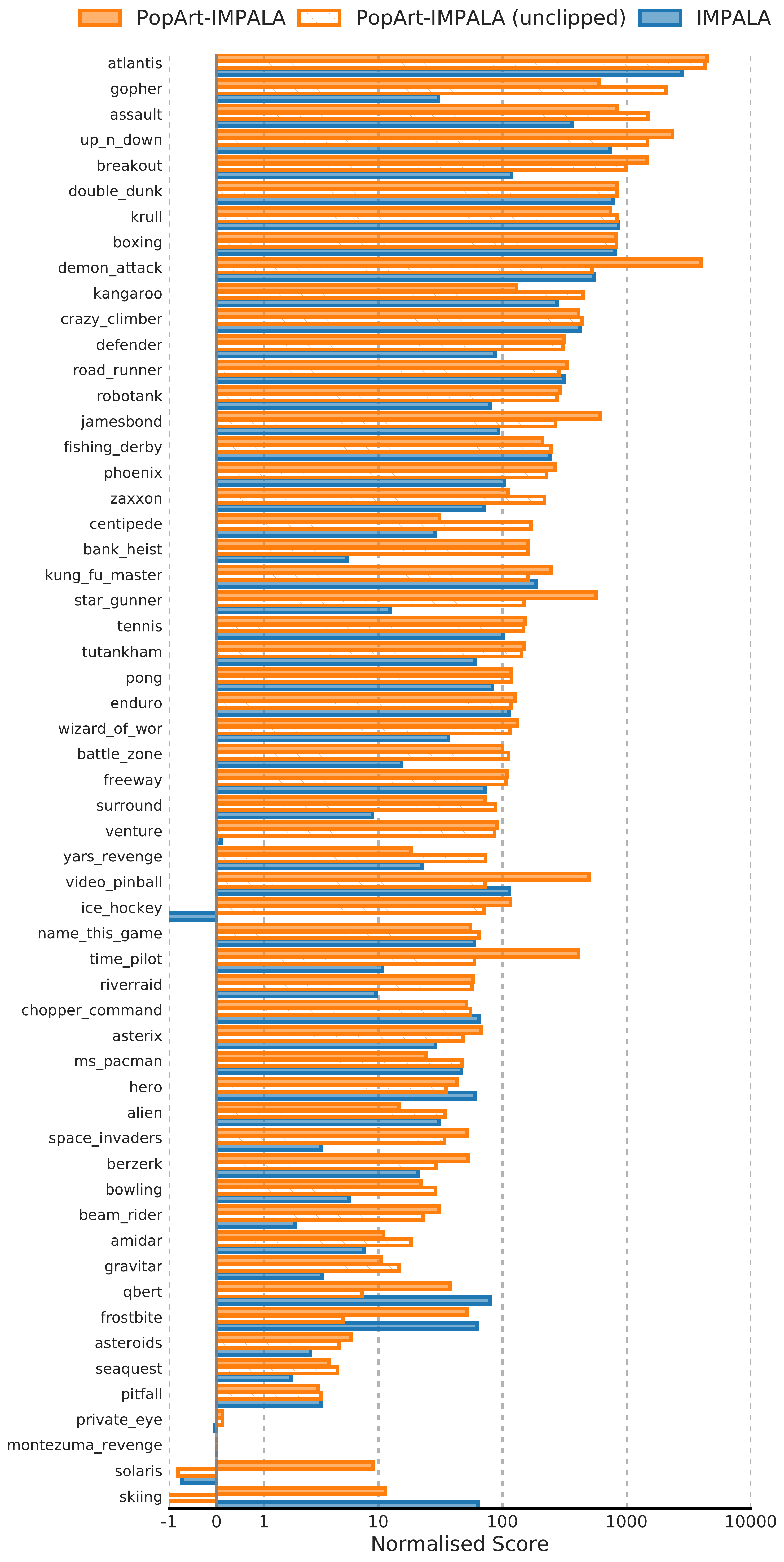}
\label{fig:stats_atari}
\end{figure}

\subsection{DmLab-30 Score breakdown}

In this section we use barplots to report the final performance of the agents on each of the levels in the DmLab-30 multi-task benchmark. In order to compute these scores we take the final trained agent and evaluate it with a frozen policy on each of the levels for 500 episodes. We perform the evaluation over a higher number of episodes (compared to Atari) because the variance of the mean episode return is typically higher in DeepMind Lab. As before, the same trained policy is evaluated on all levels, and the policy is not provided information about the task it's being evaluated on.
Also in DmLab-30 we perform a three-way comparison. We compare PopArt-IMPALA to our improved IMPALA baseline, and, for completeness, to the original paper's IMPALA. 

\begin{figure}[h!]
\centering
\caption{\textbf{DmLab-30 breakdown}: human normalised score for the original paper's IMPALA, our improved IMPALA baseline, and PopArt-IMPALA, as measured at the end of training, broken down for the 30 tasks; they all used 8 instances for population based training.}
\includegraphics[width=0.87\linewidth]{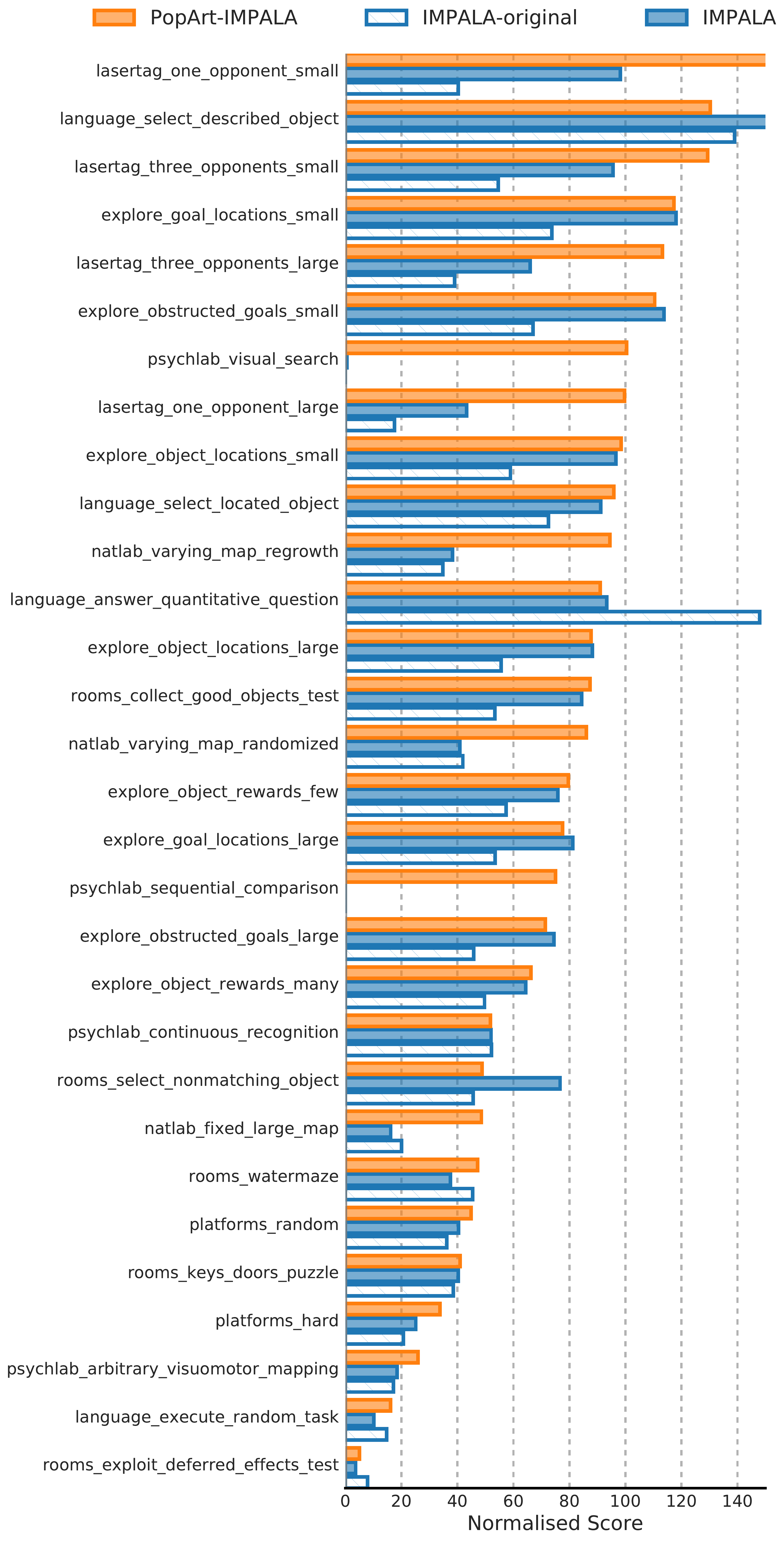}
\label{fig:stats_dmlab}
\end{figure}

\newpage

\subsection{DeepMind Lab action discretisation}
\vspace{19pt}
DeepMind Lab's native action space is a 7-dimensional continuous space, whose dimensions correspond to rotating horizontally/vertically, strafing left/right, moving forward/backward, tagging, crouching, and jumping. 

Despite the native action space being continuous, previous work on this platform has however typically relied on a coarse discretisation of the action space. We therefore follow the same approach also in our experiments.

Below we list the discretisations used by the agents considered in our experiments. This includes the discretisation used by IMPALA, as well as the one we introduce in this paper in order to unlock some levels in DmLab-30 which just can't be solved under the original IMPALA discretisation.

\vspace{15pt}

\begin{table}[h!]
\caption{Action discretisation used by IMPALA: we report below the discretisation of DeepMind Lab's action space, as used by the original IMPALA agent in Espeholt et al.}

\begin{center}
\vspace{10pt}
\begin{tabular}{ l c }

Action & Native DmLab Action  \\

Forward (FW) & \texttt{[\ \ 0, 0,\ \ 0,\ \ 1, 0, 0, 0]} \\
Backward (BW) & \texttt{[\ \ 0, 0,\ \ 0, -1, 0, 0, 0]} \\
Strafe Left & \texttt{[\ \ 0, 0, -1,\ \ 0, 0, 0, 0]} \\
Strafe Right & \texttt{[\ \ 0, 0,\ \ 1,\ \ 0, 0, 0, 0]} \\
Look Left (LL) & \texttt{[-20, 0,\ \ 0,\ \ 0, 0, 0, 0]} \\
Look Right (LR) & \texttt{[ 20, 0,\ \ 0,\ \ 0, 0, 0, 0]} \\
FW + LL & \texttt{[-20, 0,\ \ 0,\ \ 1, 0, 0, 0]} \\
FW + LR & \texttt{[ 20, 0,\ \ 0,\ \ 1, 0, 0, 0]} \\
Fire & \texttt{[\ \ 0, 0,\ \ 0,\ \ 0, 1, 0, 0]} \\

\end{tabular}
\label{tab:action_sets}
\end{center}
\end{table}

\vspace{-5pt}

\begin{table}[h!]
\centering
\caption{Action discretisation of DeepMind Lab's action space, as used by our version of IMPALA and by PopArt-IMPALA.}
\vspace{10pt}
\begin{tabular}{ l c c }

Action            & Native DmLab Action\\

FW & \texttt{[\ \ 0, 0,\ \ 0,\ \ 1, 0, 0, 0]} \\
BW & \texttt{[\ \ 0, 0,\ \ 0, -1, 0, 0, 0]} \\
Strafe Left & \texttt{[\ \ 0, 0, -1,\ \ 0, 0, 0, 0]} \\
Strafe Right & \texttt{[\ \ 0, 0,\ \ 1,\ \ 0, 0, 0, 0]} \\
Small LL  & \texttt{[-10, 0,\ \ 0,\ \ 0, 0, 0, 0]} \\
Small LR  & \texttt{[ 10, 0,\ \ 0,\ \ 0, 0, 0, 0]} \\
Large LL  & \texttt{[-60, 0,\ \ 0,\ \ 0, 0, 0, 0]} \\
Large LR  & \texttt{[ 60, 0,\ \ 0,\ \ 0, 0, 0, 0]} \\
Look Down & \texttt{[ 0, 10,\ \ 0,\ \ 0, 0, 0, 0]} \\
Look Up & \texttt{[ 0,-10,\ \ 0,\ \ 0, 0, 0, 0]} \\
FW + Small LL & \texttt{[-10, 0,\ \ 0,\ \ 1, 0, 0, 0]} \\
FW + Small LR & \texttt{[ 10, 0,\ \ 0,\ \ 1, 0, 0, 0]} \\
FW + Large LL & \texttt{[-60, 0,\ \ 0,\ \ 1, 0, 0, 0]} \\
FW + Large LR & \texttt{[ 60, 0,\ \ 0,\ \ 1, 0, 0, 0]} \\
Fire & \texttt{[\ \ 0, 0,\ \ 0,\ \ 0, 1, 0, 0]} \\

\end{tabular}
\end{table}
\vspace{10pt}

\subsection{Fixed Hyperparameters}

\begin{table}[h!]
\centering
\caption{PopArt specific hyperparameters: these are held fixed during training and were only very lightly tuned. The lower bound is used to avoid numerical issues when rewards are extremely sparse.}
\vspace{10pt}
\begin{tabular}{ l  c c }

Hyperparameter           & value \\

Statistics learning rate & \texttt{0.0003} \\
Scale lower bound & \texttt{0.0001} \\
Scale upper bound & \texttt{1e6} \\

\end{tabular}
\end{table}

\begin{table}[h!]
\centering
\caption{DeepMind Lab preprocessing. As in previous work on DeepMind Lab, we render the observation with a resolution of [72, 96], as well as use 4 action repeats. We also employ the optimistic asymmetric rescaling (OAR) of rewards, that was introduced in Espeholt et al. for exploration.}
\vspace{10pt}
\begin{tabular}{ l c c }

Hyperparameter           & value \\

Image Height & \texttt{72} \\
Image Width & \texttt{96} \\
Number of action repeats & \texttt{4} \\
Reward Rescaling & \texttt{-0.3min(tanh(r),0)+} \\
 & \texttt{5max(tanh(r),0)} \\

\end{tabular}
\end{table}

\begin{table}[h!]
\centering
\caption{Atari preprocessing. The standard Atari-preprocessing is used in the Atari experiments. Since the introduction of DQN these setting have become a standard practice when training deep RL agent on Atari. Note however, that we report experiments training agents both with and without reward clipping.}
\vspace{10pt}
\begin{tabular}{ l c c }

Hyperparameter           & value \\

Image Height & \texttt{84} \\
Image Width & \texttt{84} \\
Grey scaling & \texttt{True} \\
Max-pooling 2 consecutive frames & \texttt{True} \\
Frame Stacking & \texttt{4} \\
End of episode on life loss & \texttt{True} \\
Reward Clipping (if used) & \texttt{ [-1, 1]} \\
Number of action repeats & \texttt{4} \\

\end{tabular}
\end{table}

\begin{table}[h!]
\centering
\caption{Other agent hyperparameters: These hyperparameters are the same used by Espeholt et al.}
\vspace{10pt}
\begin{tabular}{ l c c }

Hyperparameter           & value \\

Unroll length & \texttt{20 (Atari), 100 (DmLab)}\\
Discount $\gamma$ & \texttt{0.99} \\
Baseline loss weight $\gamma$ & \texttt{0.5} \\
Batch size & \texttt{32} \\
Optimiser & \texttt{RMSProp} \\
RMSProp momentum & \texttt{0.} \\

\end{tabular}
\end{table}

\subsection{Network Architecture}

\begin{table}[h!]
\centering
\caption{Network hyperparameters. The network architecture is described in details in Espeholt et al., For completeness, we also report in the Table below the complete specification of the network. Convolutional layers are specified according to the pattern (num\_layers, kernel\_size, stride).}
\vspace{10pt}
\begin{tabular}{ l c c }

Hyperparameter           & value \\

Convolutional Stack & \\
- Number of sections & \texttt{3} \\
- Channels per section & \texttt{[16, 32, 32]} \\
- Activation Function & \texttt{ReLU} \\
ResNet section \\
- Conv & \texttt{1 / 3x3 / 1)} \\
- Max-Pool & \texttt{1 / 3x3 / 2} \\
- Conv & \texttt{2 / 3x3 / 1} \\
- Skip & \texttt{Identity} \\
- Conv & \texttt{2 / 3x3 / 1} \\
- Skip & \texttt{Identity} \\
Language preprocessing & \\
- Word embeddings & \texttt{20} \\
- Sentence embedding  & \texttt{LSTM / 64} \\
Fully connected layer & \texttt{256} \\
LSTM (DmLab-only) & \texttt{256}\\
Network Heads & \\
- Value & \texttt{Linear} \\
- Policy & \texttt{Linear+softmax} \\

\end{tabular}
\end{table}

\subsection{Population Based Training}

\begin{table}[h!]
\centering
\caption{Population Based Training: we use PBT for tuning hyper-parameters, as described in Espeholt et al., with population size and fitness function as defined below.}
\vspace{10pt}
\begin{tabular}{ l c c }

Hyperparameter           & value \\

Population Size (Atari) & \texttt{24} \\
Population Size (DmLab) & \texttt{8} \\
Fitness & \texttt{Mean capped} \\
& \texttt{human normalised} \\
& \texttt{score (cap=100)} \\

\end{tabular}
\end{table}
\vspace{-15pt}
\begin{table}[h!]
\centering
\caption{hyperparameters tuned with population based training are listed below: note that these are the same used by all baseline agents we compare to, to ensure fair comparisons.}
\vspace{10pt}
\begin{tabular}{ l c c }

Hyperparameter           & distribution \\

Entropy cost & \texttt{Log-uniform on}\\
 & \texttt{[5e-5, 1e-2]}\\
Learning rate & \texttt{Log-uniform on}\\
 & \texttt{[5e-6, 5e-3]}\\
RMSProp epsilon & \texttt{Categorical on}\\
 & \texttt{[1e-1, 1e-3, 1e-5, 1e-7]}\\
Max Grad Norm & \texttt{Uniform on}\\
 & \texttt{[10, 100]}\\

\end{tabular}
\end{table}

\end{document}